\documentclass[letterpaper]{article} 
\usepackage{aaai2026}  
\usepackage{times}  
\usepackage{helvet}  
\usepackage{courier}  
\usepackage[hyphens]{url}  
\usepackage{graphicx} 
\usepackage{natbib}  
\usepackage{caption} 
\usepackage{algorithm}
\usepackage{algorithmic}

\usepackage{amsmath}
\usepackage{booktabs}
\usepackage{multirow}
\usepackage{amssymb}
\usepackage{pifont}
\usepackage{nameref}

\newcommand{\cmark}{\ding{51}}
\newcommand{\xmark}{\ding{55}}
\newcommand{\upar}{\(\uparrow\)}    
\newcommand{\downar}{\(\downarrow\)}
\usepackage[frozencache,cachedir=minted-cache]{minted}
\usepackage{newfloat}
\usepackage{listings}
\DeclareCaptionStyle{ruled}{labelfont=normalfont,labelsep=colon,strut=off} 
\floatstyle{ruled}
\newfloat{listing}{tb}{lst}{}
\floatname{listing}{Listing}
\nocopyright

\title{AutoLink: Autonomous Schema Exploration and Expansion for Scalable Schema Linking in Text-to-SQL at Scale}
\author{
    Ziyang Wang\textsuperscript{\rm 1}\equalcontrib,
    Yuanlei Zheng\textsuperscript{\rm 1}\equalcontrib,
    Zhenbiao Cao\textsuperscript{\rm 1},
    Xiaojin Zhang\textsuperscript{\rm 2},
    Zhongyu Wei\textsuperscript{\rm 3},\\
    Pei Fu\textsuperscript{\rm 4},
    Zhenbo Luo\textsuperscript{\rm 4},
    Wei Chen\textsuperscript{\rm 1}\thanks{Corresponding author.},
    Xiang Bai\textsuperscript{\rm 1}
}
\affiliations{
    \textsuperscript{\rm 1}School of Software Engineering, Huazhong University of Science and Technology\\
    \textsuperscript{\rm 2}School of Computer Science and Technology, Huazhong University of Science and Technology\\
    \textsuperscript{\rm 3}School of Data Science, Fudan University\\
    \textsuperscript{\rm 4}MiLM Plus, Xiaomi Inc.\\
    wangziyang@hust.edu.cn, lemuria\_chen@hust.edu.cn
}

\usepackage{bibentry}

\begin{document}

\maketitle

\begin{abstract}

For industrial-scale text-to-SQL, supplying the entire database schema to Large Language Models (LLMs) is impractical due to context window limits and irrelevant noise. Schema linking, which filters the schema to a relevant subset, is therefore critical. However, existing methods incur prohibitive costs, struggle to trade off recall and noise, and scale poorly to large databases. We present \textbf{AutoLink}, an autonomous agent framework that reformulates schema linking as an iterative, agent-driven process. Guided by an LLM, AutoLink dynamically explores and expands the linked schema subset, progressively identifying necessary schema components without inputting the full database schema. Our experiments demonstrate AutoLink's superior performance, achieving state-of-the-art strict schema linking recall of \textbf{97.4\%} on Bird-Dev and \textbf{91.2\%} on Spider-2.0-Lite, with competitive execution accuracy, i.e., \textbf{68.7\%} EX on Bird-Dev (better than CHESS) and \textbf{34.9\%} EX on Spider-2.0-Lite (ranking 2nd on the official leaderboard). Crucially, AutoLink exhibits \textbf{exceptional scalability}, \textbf{maintaining high recall}, \textbf{efficient token consumption}, and \textbf{robust execution accuracy} on large schemas (e.g., over 3,000 columns) where existing methods severely degrade—making it a highly scalable, high-recall schema-linking solution for industrial text-to-SQL systems.

\end{abstract}

\begin{links}
    \link{Code}{https://github.com/wzy416/AutoLink}
\end{links}

\section{Introduction}

Text-to-SQL translates natural-language questions into executable SQL over a given database schema, lowering the barrier for non-experts~\cite{katsogiannis2023survey,DAWN}. Recent systems rely on autoregressive LLMs: the question and a structured schema representation (e.g., \emph{table/column names}, \emph{descriptions}, and \emph{primary/foreign keys}) are fed into the model, which then generates the SQL sequence~\cite{shi2024survey,hong2025nextgenerationdatabaseinterfacessurvey}. However, in large industrial databases, supplying the entire schema $S_{\text{full}}$ introduces \textbf{substantial noise from irrelevant elements} and \textbf{the risk of exceeding context window limits}, hindering correct SQL generation~\cite{spider2.0}.

To address these limitations, \textbf{\emph{Schema Linking}} emerges as a critical sub-task. Schema linking aims to identify a relevant subset of schema elements ($S_{\text{linked}} \subset S_{\text{full}}$) that are necessary to answer the user's question, thereby reducing the input context and mitigating noise for the subsequent SQL generation module~\cite{RAT-SQL}. The effectiveness of schema linking is often measured by its \textbf{strict recall rate (SRR)}~\cite{RSL-SQL}, defined as the proportion of ground-truth schema elements that are successfully included in $S_{\text{linked}}$. A high SRR is paramount, as missing essential schema elements directly limits the upper bound of SQL generation accuracy.

Existing schema linking methods include discriminative scoring of individual tables/columns, e.g., cross-encoders~\cite{resdsql} or LLM-based scoring~\cite{chess}, whole-schema reasoning and selection, e.g., full-schema prompting and backward/two-stage pipelines~\cite{MCS-SQL,SQL-TO-SCHEMA}, graph-based modeling of question–schema structure~\cite{li2023graphix}, and dual-encoder retrieval as a front-end to accelerate candidate generation; however, these routes share scalability drawbacks in industrial-scale settings because \textbf{computation} and \textbf{context windows} become bottlenecks, and achieving high SRR typically requires large candidate sets that reintroduce noise and inflate token usage, undermining the goal of schema linking.

To overcome these challenges, we introduce \textbf{AutoLink}, a novel schema linking method that redefines the problem as an interactive, sequential discovery process. Inspired by human database engineers' exploratory workflow, AutoLink employs a large language model powered autonomous agent to dynamically identify and progressively build the relevant schema subset for a natural language question. Crucially, the agent operates without requiring input of the entire database schema. It achieves this by interacting with two specialized environments: one for \emph{direct database exploration} and another for \emph{efficient semantic schema search}. Through a multi-turn dialogue, the agent strategically utilizes a diverse set of actions, including \emph{schema exploration}, \emph{semantic retrieval}, \emph{schema verification}, and \emph{schema expansion}—to iteratively refine the linked schema. This iterative and exploratory approach enables AutoLink to accurately pinpoint necessary schema elements while effectively filtering out irrelevant information, providing a highly recall schema for subsequent SQL generation.

We rigorously evaluate \textbf{AutoLink} on challenging, large-scale text-to-SQL benchmarks, Spider 2.0-Lite and Bird dataset. Our experiments demonstrate that AutoLink significantly advances schema linking, achieving state-of-the-art SRR while maintaining superior token efficiency. Notably, AutoLink achieves an SRR of \textbf{91.2\%} on Spider 2.0-Lite, dramatically outperforming baselines and maintaining high recall. This robust scalability is coupled with the lowest average token usage across all database scales, stemming from its iterative schema expansion. Furthermore, our findings reveal a critical correlation: higher SRR directly translates to improved SQL execution accuracy (EX) in downstream tasks. AutoLink consistently achieves competitive EX (e.g., \textbf{34.92\%} on Spider 2.0-Lite and \textbf{68.71\%} on Bird), often with superior token efficiency, such as requiring less than half the tokens compared to leading approaches on Spider 2.0-Lite. Ablation studies further confirm the critical contribution of each interactive action—particularly semantic retrieval—to AutoLink's robust performance and its ability to effectively handle complex, real-world industrial databases.

\section{Related Work}

Prior work on schema linking falls into two broad families. \textbf{\textit{Element-level schema linking}} scores individual tables or columns for a given question, typically via cross-encoders or LLM-based rankers; representative methods include RESDSQL~\cite{resdsql}, CodeS~\cite{CODES}, and CHESS~\cite{chess}. \textbf{\textit{Database-level schema linking}} reasons over the entire schema and the question, with three common lines: \textbf{(i)} \textbf{full-schema prompting and multi-prompt aggregation}—DIN-SQL~\cite{DIN-SQL}, MCS-SQL~\cite{MCS-SQL}, C3~\cite{C3}, DAIL-SQL~\cite{gao2023text}, E-SQL~\cite{caferouglu2024sql}, Distillery-SQL~\cite{maamari2024death}, Solid-SQL~\cite{liu2024solid}, TA-SQL~\cite{TA-SQL}, Reasoning-SQL~\cite{pourreza2025reasoning}, SQL-R1~\cite{ma2025sql}; \textbf{(ii)} \textbf{backward schema linking}—SQL-to-Schema~\cite{SQL-TO-SCHEMA}, RSL-SQL~\cite{RSL-SQL}; and \textbf{(iii)} \textbf{graph-based} approaches—RAT-SQL~\cite{RAT-SQL}, LGESQL~\cite{cao2021lgesql}, SADGA~\cite{sadga}, S$^2$SQL~\cite{hui2022s}, ISESL-SQL~\cite{semantic}, ShadowGNN~\cite{shadowgnn}, SchemaGraphSQL~\cite{safdarian2025schemagraphsql}.

Building on the above taxonomy, deploying schema linking at industrial scale exposes three recurring bottlenecks. First, \textbf{context limitations}: database-level methods often exceed LLM context windows and inflate computation on large schemas. Second, \textbf{inefficiency}: element-level methods require $O(|S|)$ scoring passes, which become impractical as $|S|$ grows. Third, a \textbf{recall–noise trade-off}: dual-encoder retrievers achieve high strict recall mainly by returning many candidates, reintroducing irrelevant schema elements and negating token savings.

\begin{figure*}[htb]
    \centering
    \includegraphics[width=\linewidth]{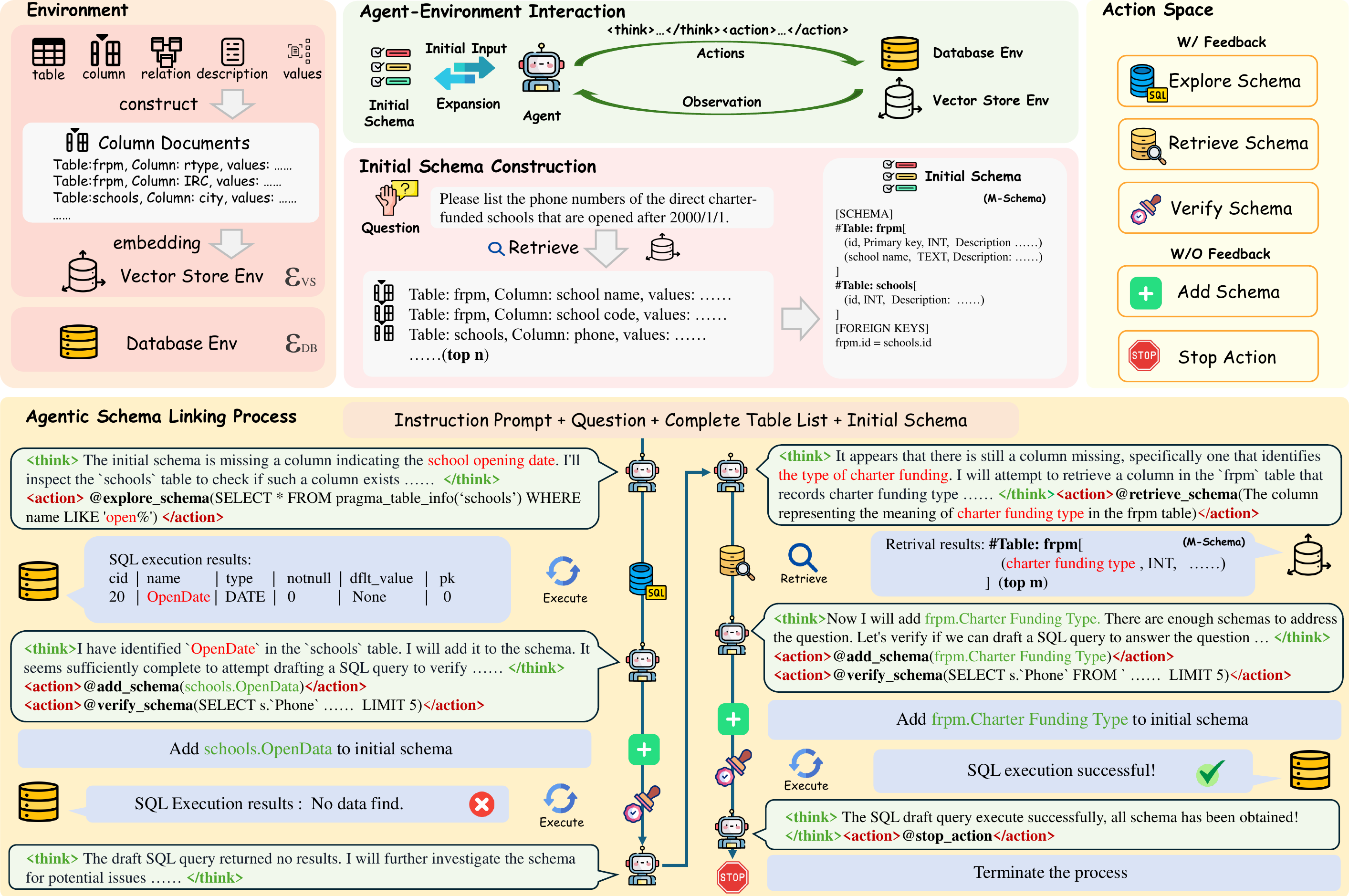}
    \caption{Overview of AutoLink: iterative interactive agentic schema linking framework.}
    \label{overview}
\end{figure*}

\section{Method}

To address the limitations of traditional schema linking methods in large industrial databases, we draw inspiration from the \textbf{exploratory} and \textbf{interactive} workflows of human database engineers. Instead of memorizing an unfamiliar database, they typically locate information through multi-step targeted SQL queries and semantic search. Inspired by this pragmatic approach, in this paper, we reframe schema linking as an interactive, sequential decision-making problem. We propose \textbf{AutoLink}, a novel schema linking method built around an autonomous agent powered by a Large Language Model (LLM), tasked with dynamically exploring and expanding the linked schema subset (\textbf{$S_{\text{linked}}$}) for given natural language question ($Q$) \textbf{without ever seeing the full database schema ($S_{\text{full}}$}). This process is modeled as the agent, guided by an LLM policy ($\pi$), executing a sequence of actions within external environments ($\mathcal{E}$), with the aim of maximizing the strict recall of ground-truth schema elements based on the interaction trajectory. The overall framework of our proposed method is illustrated in Figure~\ref{overview}. 

\subsection{External Environment $\mathcal{E}$}

To enable agent probing, a pre-built environment ($\mathcal{E}$) is provided for each unique database, comprising two distinct, complementary components: the \emph{Database Environment} $\mathcal{E}_{\text{DB}}$ and the \emph{Schema Vector Store Environment} $\mathcal{E}_{\text{VS}}$.

\noindent\textbf{Env1: Database Environment ($\mathcal{E}_{\text{DB}}$)} \quad This environment is the live database itself, providing a direct interface for schema and data exploration via SQL execution. It is defined as a function that maps an input SQL query to a formatted execution result:
\begin{equation}
\mathcal{E}_{\text{DB}}: \text{SQL} \to R_{\text{SQL}},
\end{equation}
the output $R_{\text{SQL}}$ is a structured textual response designed for exploratory operations. The $R_{\text{SQL}}$ output dynamically provides either execution results (truncated if the number of rows exceeds 5) for successful query, or specific error / timeout messages for failures, enabling clear and actionable agent feedback. 

\noindent\textbf{Env2: Schema Vector Store Environment ($\mathcal{E}_{\text{VS}}$)} \quad To facilitate efficient semantic search, we construct a vector database of all columns in the schema. For each column $c_i \in S_{\text{full}}$, we create a textual document by concatenating its core metadata: the column name, parent table's name, data type, and description (if available). We then use a powerful text encoder, i.e., bge-large-en-v1.5~\cite{chen2024bge}, to compute a dense vector representation for each column document, which is indexed into a vector store $\mathcal{V}$.

This environment is designed to bridge the semantic gap between a natural language concept and concrete schema elements. Its function is defined as:
\begin{equation}
\mathcal{E}_{\text{VS}}: (q_{\text{nl}}, K, \mathcal{C}_{\text{excl}}) \xrightarrow{\text{retrieve top-}K \text{ columns}} S_{\text{subset}},
\end{equation}
the environment takes a natural language query $q_{\text{nl}}$, a retrieval count $K$, and a set of already retrieved columns $\mathcal{C}_{\text{excl}}$ to avoid redundancy. It performs an Approximate Nearest Neighbor (ANN) search~\cite{liu2004investigation} within the vector space of columns not in $\mathcal{C}_{\text{excl}}$. While the search identifies the top-$K$ most relevant columns, the output $S_{\text{subset}}$ is a \textbf{fully structured schema snippet}. This snippet is constructed by gathering all metadata for the retrieved columns, grouping them by their parent tables, and formatting them in a human-readable \textbf{M-Schema} style~\cite{gao2024preview} that includes data types, keys, and sample values.

We provide \textbf{more details on the setup and construction of these two environments} in Appendix \nameref{Details of Environment Construction}.

\subsection{Agent-Environment Interaction}

Our LLM agent, acting as the decision-making policy $\pi$, engages the environment $\mathcal{E}$ in a multi-turn dialogue, driven by the interaction history $H$. This policy is entirely \textbf{prompt-based (training-free)} and the complete prompt is provided in Appendix \nameref{Prompt_Templates}. The objective is to progressively add potential relevant schema elements for a given user question, obtaining the final linked schema $S_{\text{linked}} \subset S_{\text{full}}$. The process commences with an initial context $H_0$. To effectively initiate the multi-turn exploration, the agent's first-turn input $H_0$ is meticulously constructed by concatenating four critical components: 

\textbf{1) Instruction Prompt ($I$)} \quad High-level instructions defining the agent's goal and available actions.

\textbf{2) User Question ($Q$)} \quad The original user query.

\textbf{3) Complete Table List ($T$)} \quad All table names in the database (excluding any columns, the total number of tables is typically manageable, e.g., $<100$), which provides essential structural context to the agent (the agent needs to know which tables it can query). 

\textbf{4) Initial Schema ($S_{\text{linked}}^{(0)}$)} \quad This component is generated by a \textbf{single, non-agent-driven call} to the schema vector store environment $\mathcal{E}_{\text{VS}}$. We use the original user question $Q$ as query to retrieve an initial set of candidate columns, creating the initial schema $S_{\text{linked}}^{(0)} = \mathcal{E}_{\text{VS}}(Q, n, \emptyset)$, where \textbf{$n$ is a relatively large hyperparameter} (e.g., 50 or 100) to ensure a broad, albeit potentially incomplete, initial set of schema elements highly relevant to the user's question. \textbf{The trade-off between recall and noise will be examined in the experimental section}. This provides the agent with crucial structural context beyond just table names, facilitating subsequent decision-making. $S_{\text{linked}}$ is also initialized directly with $S_{\text{linked}}^{(0)}$, i.e., $S_{\text{linked}} = S_{\text{linked}}^{(0)}$, and will then be updated throughout the interaction. 

In \textbf{each subsequent turn} $t$, the agent generates a reasoning trace $\theta_t$ (enclosed within \verb|<think></think>| tags) and a set of actions $A_t$ (within \verb|<actions></actions>| tags) based on the current history $H_t$:
\begin{equation}
(\theta_t, A_t) = \pi(H_t).
\end{equation}

The environment $\mathcal{E}$ then executes these actions, yielding an observation $O_t$:
\begin{equation}
O_t = \mathcal{E}(A_t).
\end{equation}

This resulting triplet $(\theta_t, A_t, O_t)$, representing a full turn of interaction, is appended to the history to form $H_{t+1}$. This loop continues until the agent terminates the process. 

\subsection{Action Space and Agentic Schema Linking}

The agent's action space $\mathcal{A}$ provides a set of specialized tools for exploration, verification, and state management. These actions are divided into two categories based on their interaction with the environment.

\noindent\textbf{Actions with Feedback} \quad These are the agent's primary tools for gathering new information from the environment.

\noindent\textbf{1. @explore\_schema} \quad This action aims to interact with the Database Environment ($\mathcal{E}_{\text{DB}}$) by executing exploratory SQL queries, primarily targeting schema metadata or small samples of data, \textbf{rather than directly answering the user's question}. For example, the agent can query all columns that contain certain characters (e.g., \emph{id} or \emph{name}) with fuzzy matching, column descriptions and sample column values for certain columns, or examine a table's primary or foreign keys, and other structural metadata, etc.

\noindent\textbf{2. @retrieve\_schema} \quad This action directly explores missing schema elements within the Schema Vector Store Environment ($\mathcal{E}_{\text{VS}}$). Unlike simple user-question rewrite like CHESS~\cite{chess}, it provides a strong signal for missing schema search. Leveraging the user's natural language question, known table names, and the currently incomplete schema, the agent can directly \textbf{infer required missing \emph{column names}, \emph{descriptions} or \emph{concepts} as new query action}. This approach is especially powerful for high-level or ambiguous user queries, narrowing the semantic search space and enabling the discovery of specific, relevant columns beyond simple rephrasing. 

Specifically, this action generates a set of retrieved candidate schema elements as: 
$S_{\text{retrieved}} = \mathcal{E}_{\text{VS}}(q, m, \mathcal{C}_{\text{excl}})$, where \textbf{$m$ is a relatively small number} (e.g., 3 or 5) since this action is a targeted retrieval. Previously retrieved columns (from the initial schema or prior \textbf{@retrieve\_schema} calls) are excluded from the search space. 

\noindent\textbf{3. @verify\_schema} \quad This action interacts with $\mathcal{E}_{\text{DB}}$ and functions as a holistic hypothesis test by attempting to \textbf{execute a full SQL query specifically designed to answer the user's question}. The primary purpose is \textbf{not to generate the final query result}, but to check for schema sufficiency. A successful execution may indicate completeness, while an error provides a strong, direct signal about which specific schema elements are still missing.  

\noindent\textbf{Actions without Feedback} \quad These actions manage the agent's internal state and control the workflow.

\noindent\textbf{4. @add\_schema} \quad This action serves as the agent's mechanism for explicitly committing newly discovered relevant schema elements and updating the linked schema $S_{\text{linked}}$, which we call \textbf{Schema Expansion}. After actions with feedback yields new, relevant schema elements, the agent can adopt this action. To improve schema linking recall, we particularly encourage the agent to add schema returned by the \textbf{@retrieve\_schema} action.

The agent's output for this action specifies the schema elements to be added, presented as a semicolon-separated list of \emph{table\_name.column\_name} strings within the \emph{add\_schema()}. Upon execution, the system processes these identifiers by collecting their complete meta-information. Let $S_{\text{added}}$ denote the set of these fully described schema elements. $S_{\text{linked}}$ is then updated by merging these newly added schema elements:
\begin{equation}
S_{\text{linked}} \leftarrow S_{\text{linked}} \cup S_{\text{added}}.
\end{equation}
\textbf{It is important to note that while the agent can output one or more actions per turn, @add\_schema cannot be the sole action}, since outputting it alone would leave the agent without environmental feedback for subsequent turns. Therefore, the instruction prompt explicitly requires \textbf{@add\_schema} to always be paired with at least one feedback-providing action or \textbf{@stop\_action}.

\noindent\textbf{5. @stop\_action} \quad  This action terminates the multi-turn interaction process. The agent uses this action when, based on its analysis of the dialogue history (including initial schema and expanded schema elements through previous \textbf{@add\_schema} actions), it determines that enough information has been gathered to answer the user's question. This decision primarily relies on the agent's contextual understanding and significantly influenced by the outcomes of the \textbf{@verify\_schema} action. Additionally, the process is also terminated if the number of interaction turns exceeds a predefined maximum (\textbf{10 turns} in this paper), serving as a safeguard against endless loops.

\subsection{SQL Generation}

With the final linked schema $S_{\text{linked}}$ obtained through our iterative, agent-driven process, the subsequent step is \textbf{SQL generation}. It is important to note that SQL generation itself is \textbf{not the core focus} of this paper, and we leverage existing techniques for this phase. We employ an existing LLM as the SQL generation policy ($\pi_{\text{SQL}}$). Specifically, given the user's original natural language question $Q$ and final linked schema $S_{\text{linked}}$, the policy adopts a self-consistency strategy~\cite{wang2022self} to sample multiple SQL candidates, then performs syntactic correction~\cite{chess,chase} and majority voting~\cite{REFORCE} to derive the final SQL statement. Please refer to the Appendix \nameref{appendix:sql-generation}. 

\begin{table*}[htb]
\setlength{\tabcolsep}{1.2mm}{
\begin{tabular}{lcccccccccc}
\toprule
& & \multicolumn{3}{c}{Bird Dev} & \multicolumn{6}{c}{Spider2.0-Lite} \\ 
\cmidrule(lr){3-5} \cmidrule(lr){6-11} \multirow{-2}{*}{Method} & \multirow{-2}{*}{Full Input} & SRR$^\uparrow$   & $\bar{C}^\downarrow$     & Avg. Tokens$^\downarrow$  & SRR$_{all}^\uparrow$  & SRR$_{bq}^\uparrow$   & SRR$_{sf}^\uparrow$   & SRR$_{local}^\uparrow$   & $\bar{C}^\downarrow$      & Avg. Tokens$^\downarrow$ \\ \midrule
DE-SL (BGE-Large) & \cmark & 35.5  & 35.7  & -- & 43.6 & 28.2  & 57.1  & 87.5  & 153.8 & -- \\
CE-SL (BGE-reranker) & \cmark & 72.4  & 35.7  & -- & 57.6 & 42.3  & 72.6  & \textbf{95.8}  & 153.8  & -- \\
MCS-SQL & \cmark & 85.7  & 7.5   & 29.8K & 58.9 & 57.8  & 54.8  & 79.2  & 45.1   & 168.9K      \\
SQL-to-Schema & \cmark & 92.5  & 13.5  & 19.4K & 64.0 & 64.8  & 54.8  & 91.7  & 49.0   & 171.9K      \\
CHESS & \cmark & 89.7  & \textbf{4.5}  & -- & -- & -- & -- & --    & -- & --          \\
RSL-SQL & \cmark & 93.3  & 13.0  & 14.8K & 52.0 & 52.8  & 44.1  & 75.0    & 25.8   & 29.2K       \\
LinkAlign & \xmark & --    & --    & -- & 36.4 & 22.5  & 51.2  & 66.7  & \textbf{21.1}   & 66.7K       \\ \midrule
\textbf{AutoLink (ours)} & \xmark & \textbf{97.4}  & 35.8  & \textbf{8.0K} & \textbf{91.2} & \textbf{93.7} & \textbf{85.7} & \textbf{95.8} & 159.4 & \textbf{21.2K}       \\ \bottomrule
\end{tabular}
}
\centering
\caption{\textbf{Performance comparison of schema linking on Bird Dev and Spider 2.0-Lite}. We report the overall Strict Recall Rate (SRR$_{all}$) for both datasets. For Spider 2.0-Lite, SRR is further broken down by SQL dialect: BigQuery (SRR$_{bq}$), Snowflake (SRR$_{sf}$), and SQLite (SRR$_{local}$). $\bar{C}$ denotes the average number of columns included in the simplified schema ($S_{\text{linked}}$) after schema linking. \textbf{Full Input} indicates whether the entire database schema is provided to the LLM or if all database schemas are iterated through.}
\label{link result}
\end{table*}

\section{Experiment}

\subsection{Implementation Details}

\noindent\textbf{Experimental Setup} \quad Considering the cost-effectiveness of the DeepSeek series model, DeepSeek-V3 serves as the LLM policy $\pi$ for schema linking (AutoLink), and DeepSeek-R1 and DeepSeek-V3 are employed as the policy $\pi_{\text{SQL}}$ for the subsequent SQL generation phase. The reasoning LLM (DeepSeek-R1) is not chosen for AutoLink's policy due to observed \textbf{instruction following degradation}~\cite{fu2025scaling}, which occasionally leads to deviations from the required thought and action format. The \textbf{key hyperparameters} for our AutoLink framework include top-$n$ for initial schema retrieval (varied between 5 and 100 during experiments); top-$m$ for the \textbf{@retrieve\_schema} action, fixed at 3; and a maximum interaction turn limit, set to 10.

\noindent\textbf{Datasets} \quad Our evaluation is conducted on two distinct Text-to-SQL benchmarks: the Spider 2.0-Lite dataset~\cite{spider2.0} and the Bird-Dev dataset~\cite{BIRD}. Bird-Dev comprises 11 databases, featuring an average of 80 columns per database, and includes 1,543 complex SQL query use cases. Spider 2.0-Lite, derived from industrial applications, is designed to reflect the scale of real-world databases. It presents a greater challenge than Bird-Dev, with its databases averaging over 800 columns, more intricate SQL queries, and support for multiple SQL dialects (including BigQuery, Snowflake, and SQLite). This dataset contains 547 test cases.  

\noindent\textbf{Evaluation Metrics} \quad Model performance is evaluated using 3 primary metrics. The \textbf{Strict Recall Rate (SRR)}~\cite{RSL-SQL} measures the effectiveness of schema linking, defined as the proportion of test cases where the final simplified schema fully contains all required gold schemas. For SQL generation, we report \textbf{Execution Accuracy (EX)}, consistent with the official definitions of the Spider and BIRD benchmarks, which measures the consistency between the execution results of predicted and gold SQL queries. Lastly, for LLM-based methods, we report \textbf{Avg. Token Consumption}, representing the average total tokens (input plus output) per example. This metric serves as a key metric of computational cost, as overall inference latency is highly susceptible to external factors (e.g., API, network variability), and the time spent on SQL execution within our method is negligible compared to LLM decoding.

\noindent\textbf{Baselines} \quad To evaluate AutoLink's performance, we compare it against a diverse set of established methods across both schema linking and SQL generation tasks. For element-level schema linking, our baselines include \textbf{DE-SL}, which employs a dual-encoder (BGE-Large-v1.5) to pre-cache column documents and retrieve the top-$K$ most similar columns, and \textbf{CE-SL}, which uses a cross-encoder (BGE-reranker) to compute and rank the similarity between the user query and each column individually for top-$K$ retrieval. Additionally, \textbf{CHESS}~\cite{chess} is included, an LLM-based method that scores the user query against each column, followed by sequential table and column filtering. For database-level schema linking, we consider \textbf{MCS-SQL}~\cite{MCS-SQL}, which leverages an LLM to directly output relevant schema elements from the full schema and user query, utilizing 5-time sampling decoding to merge results. Similarly, \textbf{SQL-to-schema}~\cite{SQL-TO-SCHEMA} uses an LLM to generate an SQL statement from which involved tables and columns are parsed, also employing 5-time sampling decoding. \textbf{RSL-SQL}~\cite{RSL-SQL} is a hybrid approach combining elements of MCS-SQL and SQL-to-schema, typically with a single decoding pass. Furthermore, \textbf{LinkAlign}~\cite{LINKALIGN} stands as a baseline for its query rewriting and multi-agent discussion framework that also bypasses the need for schema element retrival.

\subsection{Main Results of Schema Linking}

As shown in Table~\ref{link result}, all models (except for \textbf{DE-SL} and \textbf{CE-SL}) are implemented using DeepSeek-V3 as the backbone to ensure fair and consistent comparison. For the hyperparameter top-$n$ for initial schema retrieval, we set 30 on Bird Dev and 100 on Spider 2.0-Lite. The experimental results demonstrate that \textbf{AutoLink} achieves a significant advantage over baseline methods in terms of SRR and average token consumption across both the Bird and Spider 2.0-Lite. On the more challenging Spider 2.0-Lite benchmark, our method achieves a \textbf{27.2\%} improvement in strict recall rate compared to the second place (\textbf{SQL-to-Schema}), while reducing the maximum token consumption by \textbf{87.7\%}. Compared with encoder-based methods (\textbf{DE-SL} and \textbf{CE-SL}), although the number of recalled columns is close, our method improves SRR by an average of approximately \textbf{40\%}. 

Compared to methods requiring full schema input, such as \textbf{MCS-SQL}, \textbf{SQL-to-Schema}, and \textbf{RSL-SQL}, we observe that while they perform acceptably on smaller datasets like Bird, their performance sharply declines on larger, more complex databases like Spider 2.0-Lite. This is due to the long context lengths and high resource consumption at scale. Although these methods can theoretically enhance SRR by increasing sampling decoding iterations, we found that the SRR growth becomes negligible and quickly saturates, particularly on the Spider 2.0-Lite. \emph{Blindly increasing decoding attempts yields minimal performance gains}, as the results across different decoding passes often exhibit low diversity. Consequently, \textbf{these methods struggle to effectively control the number of recalled columns to achieve a comparable scale to ours}. In contrast, our method consistently maintains strong performance across varying dataset scales. We demonstrate the scatter plot of recalled columns versus SRR in Appendix \nameref{appendix:more experiment results}, and our approach still \textbf{achieves superior recall rates even when recalling a similar small number of columns}. Regarding the comparison of Token consumption, we found that compared with above methods, our method significantly reduces the Token overhead. Compared with RSL-SQL, although its Token overhead is not large, its recall performance in large-scale databases is relatively low.

In contrast to methods like \textbf{CHESS} and \textbf{LinkAlign}, which \textbf{prioritize noise reduction by aiming to include only the columns strictly required by the gold SQL}, our approach differs. This noise-minimization strategy carries substantial risk, as evidenced by LinkAlign's mere 36.4\% SRR, despite its minimal average of 21.1 recalled columns. Such a low recall rate severely compromises the utility of the simplified schema for subsequent SQL generation.

\begin{table}[tbp]
\centering
\setlength{\tabcolsep}{1.3mm}{
\begin{tabular}{llcc}
\toprule
Method & Model & EX$^\uparrow$  & Avg. Tokens$^\downarrow$ \\ \midrule
\multirow{6}{*}{Spider-Agent}
  & QwQ & 11.33 & -- \\
  & GPT-4o & 13.16 & -- \\
  & DeepSeek-R1 & 13.71 & -- \\
  & o1-preview & 23.03 & -- \\
  & o3-mini & 23.40 & -- \\
  & Claude-3.7-Sonnet & 28.52 & -- \\ \midrule
CHESS & GPT-4o & 3.84 & -- \\
LinkAlign & DeepSeek-R1 & 33.09 & -- \\
$\text{RSL-SQL}^\dag$ & DeepSeek-R1 & 30.53 & \underline{50.0K} \\
$\text{REFORCE}^\dag$ & DeepSeek-R1  & 29.62  & 81.1K  \\
REFORCE & o1-preview  & 30.35  & 81.1K  \\
REFORCE & GPT-o3  & \textbf{37.84}  & 81.1K  \\ \midrule
\textbf{AutoLink} & DeepSeek-R1 & \underline{34.92} & \textbf{38.0K} \\ \bottomrule
\end{tabular}
}
\caption{Execution Accuracy (EX) comparison of different methods on the Spider 2.0-Lite dataset. $^\dag$ indicates results obtained through independent reproduction. Avg. Tokens indicates the average number of tokens comsumed to generate a SQL.}
\label{EX_RESULTS_SPIDER}
\end{table}

\subsection{Results of SQL Generation}

The comparative results in Table \ref{EX_RESULTS_SPIDER} and Table \ref{EX_RESULTS_BIRD} demonstrate that \textbf{AutoLink} achieves competitive EX on both Spider 2.0-Lite and Bird. On Spider 2.0-Lite, AutoLink attains an EX score of \textbf{34.92\%} using DeepSeek-R1, which outperforms most baseline approaches and is highly competitive with the best-performing method, ReFoRCE (37.84\% with GPT-o3). Notably, ReFoRCE generates eight candidate SQL queries per example, whereas AutoLink generates only five. This difference results in a significant advantage for AutoLink in terms of token consumption—AutoLink requires only \textbf{38.0K} tokens on average, less than half of ReFoRCE’s 81.1K. When evaluated using the same model, DeepSeek-R1, our method achieves the best performance. Similarly, on the Bird dev set, AutoLink also achieves strong results. With Gemini-1.5-Pro as the backbone model, AutoLink achieves an EX score of \textbf{68.71\%}, which is competitive with or superior to strong baselines such as \textbf{CHESS} and \textbf{RSL-SQL}.

\begin{table}[tbp]
\centering
\begin{tabular}{llcc}
\toprule
Method & Model & EX$^\uparrow$  & Avg. Tokens$^\downarrow$ \\ \midrule
MAC-SQL & GPT-4 & 59.39 & \textbf{6.8K} \\
TA-SQL & GPT-4 & 56.19 & \underline{7.3K} \\
RSL-SQL & GPT-4o & 67.21  & 7.5K \\ 
CHESS & Gemini-1.5-Pro & \underline{68.31} & 14.5K \\ \midrule
\textbf{AutoLink} & DeepSeek-v3 & 66.36 & 8.0K \\
\textbf{AutoLink} & Gemini-1.5-Pro & \textbf{68.71} & 8.0K \\ \bottomrule
\end{tabular}
\caption{Execution Accuracy (EX) comparison of different methods the Bird Dev.}
\label{EX_RESULTS_BIRD}
\end{table}

\begin{table}[tbp]
\centering
\small
\begin{tabular}{lccc}
\toprule
Method                & SRR$_{n=5}$  & SRR$_{n=50}$  & SRR$_{n=100}$  \\ \midrule
AutoLink              & \textbf{79.2} & \textbf{88.0} & \textbf{91.2} \\
- w/o Verify Schema   & 77.2$_{\downarrow2.0}$ & 86.8$_{\downarrow1.2}$ & 89.6$_{\downarrow1.6}$ \\
- w/o Explore Schema  & 76.8$_{\downarrow2.4}$ & 84.8$_{\downarrow3.2}$ & 88.8$_{\downarrow2.4}$ \\
- w/o Retrieve Schema & 72.4$_{\downarrow6.8}$ & 80.4$_{\downarrow7.6}$ & 84.5$_{\downarrow6.7}$ \\ \bottomrule
\end{tabular}
\caption{Ablation study on the impact of different schema linking actions under varying numbers of initial candidates $n$ on Spider 2.0-Lite.}
\label{ABLATION_SL}
\end{table}

\begin{figure*}[ht]
    \centering
    \includegraphics[width=\linewidth]{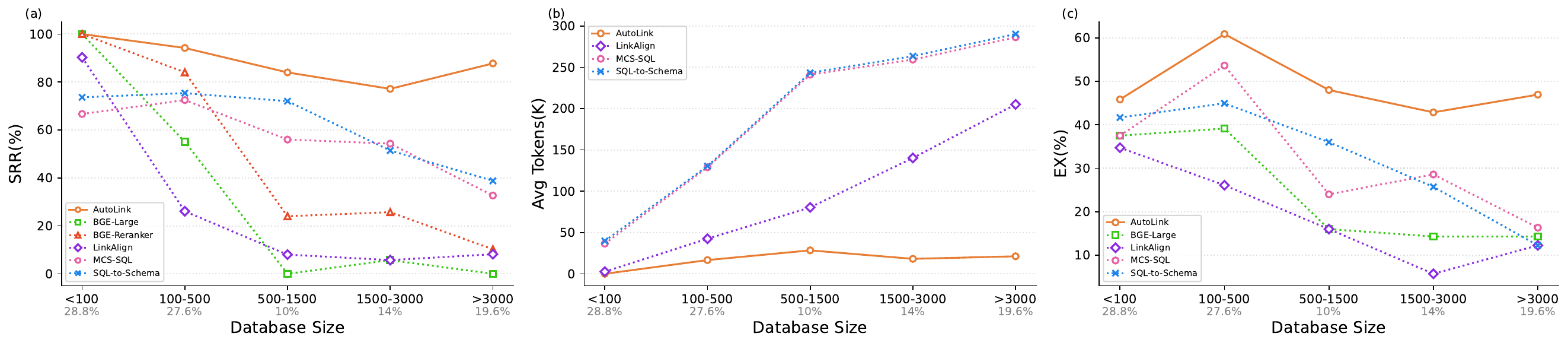}
    \caption{Scalability comparison across databases on Spider 2.0-Lite of varying sizes in terms of (a) Strict Recall Rate (SRR\upar), (b) Average Tokens Consumption (Avg. Tokens\downar), and (c) Execution Accuracy (EX\upar). Percentages below each bin indicate the proportion of databases within each size range.}
    \label{DATABASE}
\end{figure*}

\begin{figure*}[ht]
    \centering
    \includegraphics[width=\linewidth]{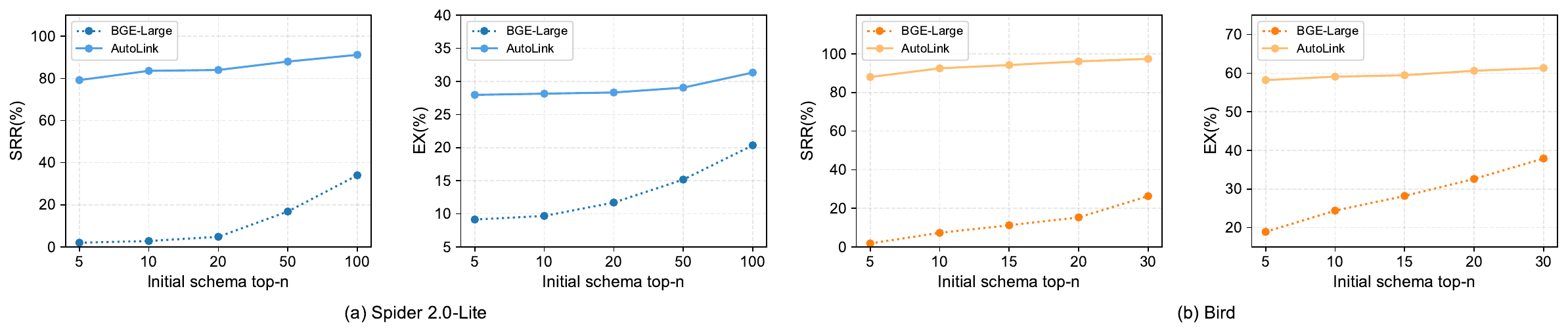}
    \caption{SRR and EX comparison of schema linking at different initial top-$n$ on Spider 2.0-Lite and Bird dev set.}
    \label{TOPK_ANA}
\end{figure*}

\subsection{Ablation Study}

The ablation results in Table~\ref{ABLATION_SL} demonstrate the importance of each core agent action for schema linking under initial retrieval sizes of top-5, top-50, and top-100. Removing any single action—\textbf{@retrieve\_schema}, \textbf{@explore\_schema}, or \textbf{@verify\_schema}—leads to a clear drop in SRR across all initial top-$n$ settings. Notably, removing schema retrieval has the largest impact, highlighting its key role in identifying relevant columns. Without this capability, the agent is much less able to anchor its reasoning in the most promising schema elements. Excluding database exploration also consistently reduces performance, underscoring its value for resolving ambiguities when facing unfamiliar databases. Although omitting verification results in a smaller decrease in SRR, its contribution is evident across all candidate sizes, reflecting the importance of continual self-assessment within the agent's reasoning process.

\subsection{Scalability Analysis of Different Database Scales} 

Figure~\ref{DATABASE} compares the SRR of various schema linking methods: \textbf{AutoLink}, \textbf{BGE-Large (DE-SL)}, \textbf{BGE-reranker (CE-SL)}, and \textbf{LinkAlign}, etc., on Spider 2.0-Lite across increasing database scales. All methods exhibit a significant decline in SRR as the database size grows, though the decrease of \textbf{AutoLink} is notably slower. In large databases exceeding 3,000 columns, the SRR of all baseline models drops below \textbf{40\%}, yet AutoLink's SRR remains near \textbf{90\%}. For token consumption ((b) in Figure~\ref{DATABASE}), all methods generally show an upward trend. AutoLink consistently demonstrates the lowest average token usage across all database sizes. This efficiency stems from AutoLink's iterative approach, which begins with a smaller schema and gradually expands it, resulting to minimal changes in token consumption across different database scales.

To ensure a fair comparison, our SQL generation method is applied to the linked schemas produced by different schema linking approaches. Regarding EX ((c) in Figure~\ref{DATABASE}), the performance of all baselines declines as schema size increases. Crucially, we observe a direct correlation: \textbf{models with higher SRR tend to exhibit higher EX}. This highlights that a higher SRR is instrumental in improving the EX of subsequent SQL generation. This is intuitively sound because if the SQL generator (i.e., the LLM), does not hallucinate, it will only utilize the schema elements provided by the schema linking step. Consequently, an incomplete schema (due to low SRR) means the generated SQL may lack necessary tables and columns from the gold SQL, making it highly probable that the generated query will be incorrect. \textbf{AutoLink} consistently achieves the highest EX across all size ranges, significantly outperforming baselines on the largest databases due to its superior SRR.

\subsection{Analysis of Hyperparameter}

As shown in Figure~\ref{TOPK_ANA}, \textbf{AutoLink} consistently outperforms BGE Large in SRR under various initial top-$n$ schema retrieval settings. Increasing top-$n$ improves SRR for both methods, but AutoLink’s agent-driven exploration gives it a notable advantage even \textbf{at small top-$n$ values} (e.g., top-5) by effectively identifying and incorporating missing schema components. This robustness holds across datasets of different scales, demonstrating good scalability.

For downstream SQL EX on Spider 2.0-Lite, AutoLink consistently achieves higher accuracy than BGE across all top-$n$ settings. While its performance rises with larger top-$n$, the improvement plateaus at high values, suggesting that \textbf{most critical schema elements can already be recovered with smaller candidate sets}. The advantage is most pronounced at low top-$n$, highlighting AutoLink's effectiveness when initial schema retrieval is incomplete. Additional hyperparameter analysis, including the effects of max turn and top-$m$ expansion, is provided in the Appendix \nameref{appendix:more experiment results}.

\section{Conclusion}
In this paper, we propose AutoLink, a novel framework that redefines schema linking as an adaptive, agent-driven process. By orchestrating semantic retrieval from a vector store and utilizing lightweight SQL probes, our LLM-powered agent iteratively and autonomously assembles only the schema elements truly necessary for a given query, without requiring the full database schema as input. This unified mechanism achieves state-of-the-art strict recall on Spider 2.0-Lite and Bird, significantly reduces token usage, and demonstrates exceptional scalability and robustness on large schemas with thousands of columns, thereby providing a practical and efficient foundation for industrial-scale Text-to-SQL systems.

\section{Acknowledgments}
This research was supported by \emph{National Natural Science Foundation of China} (No.62406121) and \emph{National Science Foundation of Hubei Province, China} (No.2024AFB189).

\bibliography{aaai2026}

\begin{thebibliography}{41}
\providecommand{\natexlab}[1]{#1}

\bibitem[{Caferoğlu and Özgür Ulusoy(2024)}]{caferouglu2024sql}
Caferoğlu, H.~A.; and Özgür Ulusoy. 2024.
\newblock E-SQL: Direct Schema Linking via Question Enrichment in Text-to-SQL.
\newblock arXiv:2409.16751.

\bibitem[{Cai et~al.(2021)Cai, Yuan, Xu, and Hao}]{sadga}
Cai, R.; Yuan, J.; Xu, B.; and Hao, Z. 2021.
\newblock Sadga: Structure-aware dual graph aggregation network for text-to-sql.
\newblock \emph{Advances in Neural Information Processing Systems}, 34: 7664--7676.

\bibitem[{Cao et~al.(2021)Cao, Chen, Chen, Zhao, Zhu, and Yu}]{cao2021lgesql}
Cao, R.; Chen, L.; Chen, Z.; Zhao, Y.; Zhu, S.; and Yu, K. 2021.
\newblock {LGESQL}: Line Graph Enhanced Text-to-{SQL} Model with Mixed Local and Non-Local Relations.
\newblock In Zong, C.; Xia, F.; Li, W.; and Navigli, R., eds., \emph{Proceedings of the 59th Annual Meeting of the Association for Computational Linguistics and the 11th International Joint Conference on Natural Language Processing (Volume 1: Long Papers)}, 2541--2555. Online: Association for Computational Linguistics.

\bibitem[{Cao et~al.(2024)Cao, Zheng, Fan, Zhang, Chen, and Bai}]{RSL-SQL}
Cao, Z.; Zheng, Y.; Fan, Z.; Zhang, X.; Chen, W.; and Bai, X. 2024.
\newblock RSL-SQL: Robust Schema Linking in Text-to-SQL Generation.
\newblock arXiv:2411.00073.

\bibitem[{Chen et~al.(2024)Chen, Xiao, Zhang, Luo, Lian, and Liu}]{chen2024bge}
Chen, J.; Xiao, S.; Zhang, P.; Luo, K.; Lian, D.; and Liu, Z. 2024.
\newblock {M}3-Embedding: Multi-Linguality, Multi-Functionality, Multi-Granularity Text Embeddings Through Self-Knowledge Distillation.
\newblock In Ku, L.-W.; Martins, A.; and Srikumar, V., eds., \emph{Findings of the Association for Computational Linguistics: ACL 2024}, 2318--2335. Bangkok, Thailand: Association for Computational Linguistics.

\bibitem[{Chen et~al.(2021)Chen, Chen, Zhao, Cao, Xu, Zhu, and Yu}]{shadowgnn}
Chen, Z.; Chen, L.; Zhao, Y.; Cao, R.; Xu, Z.; Zhu, S.; and Yu, K. 2021.
\newblock {S}hadow{GNN}: Graph Projection Neural Network for Text-to-{SQL} Parser.
\newblock In Toutanova, K.; Rumshisky, A.; Zettlemoyer, L.; Hakkani-Tur, D.; Beltagy, I.; Bethard, S.; Cotterell, R.; Chakraborty, T.; and Zhou, Y., eds., \emph{Proceedings of the 2021 Conference of the North American Chapter of the Association for Computational Linguistics: Human Language Technologies}, 5567--5577. Online: Association for Computational Linguistics.

\bibitem[{Deng et~al.(2025)Deng, Ramachandran, Xu, Hu, Yao, Datta, and Zhang}]{REFORCE}
Deng, M.; Ramachandran, A.; Xu, C.; Hu, L.; Yao, Z.; Datta, A.; and Zhang, H. 2025.
\newblock ReFo{RCE}: A Text-to-{SQL} Agent with Self-Refinement, Format Restriction, and Column Exploration.
\newblock In \emph{ICLR 2025 Workshop: VerifAI: AI Verification in the Wild}.

\bibitem[{Devlin et~al.(2019)Devlin, Chang, Lee, and Toutanova}]{devlin2019bert}
Devlin, J.; Chang, M.-W.; Lee, K.; and Toutanova, K. 2019.
\newblock Bert: Pre-training of deep bidirectional transformers for language understanding.
\newblock In \emph{Proceedings of the 2019 conference of the North American chapter of the association for computational linguistics: human language technologies, volume 1 (long and short papers)}, 4171--4186.

\bibitem[{Dong et~al.(2023)Dong, Zhang, Ge, Mao, Gao, lu~Chen, Lin, and Lou}]{C3}
Dong, X.; Zhang, C.; Ge, Y.; Mao, Y.; Gao, Y.; lu~Chen; Lin, J.; and Lou, D. 2023.
\newblock C3: Zero-shot Text-to-SQL with ChatGPT.
\newblock arXiv:2307.07306.

\bibitem[{Fu et~al.(2025)Fu, Gu, Li, Qu, and Cheng}]{fu2025scaling}
Fu, T.; Gu, J.; Li, Y.; Qu, X.; and Cheng, Y. 2025.
\newblock Scaling Reasoning, Losing Control: Evaluating Instruction Following in Large Reasoning Models.
\newblock arXiv:2505.14810.

\bibitem[{Gao et~al.(2024)Gao, Wang, Li, Sun, Qian, Ding, and Zhou}]{gao2023text}
Gao, D.; Wang, H.; Li, Y.; Sun, X.; Qian, Y.; Ding, B.; and Zhou, J. 2024.
\newblock Text-to-SQL Empowered by Large Language Models: A Benchmark Evaluation.
\newblock \emph{Proc. VLDB Endow.}, 17(5): 1132–1145.

\bibitem[{Gao et~al.(2025)Gao, Liu, Li, Shi, Zhu, Wang, Li, Li, Hong, Luo, Gao, Mou, and Li}]{gao2024preview}
Gao, Y.; Liu, Y.; Li, X.; Shi, X.; Zhu, Y.; Wang, Y.; Li, S.; Li, W.; Hong, Y.; Luo, Z.; Gao, J.; Mou, L.; and Li, Y. 2025.
\newblock A Preview of XiYan-SQL: A Multi-Generator Ensemble Framework for Text-to-SQL.
\newblock arXiv:2411.08599.

\bibitem[{Hong et~al.(2025)Hong, Yuan, Zhang, Chen, Dong, Huang, and Huang}]{hong2025nextgenerationdatabaseinterfacessurvey}
Hong, Z.; Yuan, Z.; Zhang, Q.; Chen, H.; Dong, J.; Huang, F.; and Huang, X. 2025.
\newblock { Next-Generation Database Interfaces: A Survey of LLM-Based Text-to-SQL }.
\newblock \emph{IEEE Transactions on Knowledge \& Data Engineering}, 37(12): 7328--7345.

\bibitem[{Hui et~al.(2022)Hui, Geng, Wang, Qin, Li, Li, Sun, and Li}]{hui2022s}
Hui, B.; Geng, R.; Wang, L.; Qin, B.; Li, Y.; Li, B.; Sun, J.; and Li, Y. 2022.
\newblock {S}$^2${SQL}: Injecting Syntax to Question-Schema Interaction Graph Encoder for Text-to-{SQL} Parsers.
\newblock In Muresan, S.; Nakov, P.; and Villavicencio, A., eds., \emph{Findings of the Association for Computational Linguistics: ACL 2022}, 1254--1262. Dublin, Ireland: Association for Computational Linguistics.

\bibitem[{Johnson, Douze, and J{\'e}gou(2019)}]{johnson2019billion}
Johnson, J.; Douze, M.; and J{\'e}gou, H. 2019.
\newblock Billion-scale similarity search with {GPUs}.
\newblock \emph{IEEE Transactions on Big Data}, 7(3): 535--547.

\bibitem[{Katsogiannis-Meimarakis and Koutrika(2023)}]{katsogiannis2023survey}
Katsogiannis-Meimarakis, G.; and Koutrika, G. 2023.
\newblock A survey on deep learning approaches for text-to-SQL.
\newblock \emph{The VLDB Journal}, 32(4): 905–936.

\bibitem[{Lee et~al.(2025)Lee, Park, Kim, and Park}]{MCS-SQL}
Lee, D.; Park, C.; Kim, J.; and Park, H. 2025.
\newblock {MCS}-{SQL}: Leveraging Multiple Prompts and Multiple-Choice Selection For Text-to-{SQL} Generation.
\newblock In Rambow, O.; Wanner, L.; Apidianaki, M.; Al-Khalifa, H.; Eugenio, B.~D.; and Schockaert, S., eds., \emph{Proceedings of the 31st International Conference on Computational Linguistics}, 337--353. Abu Dhabi, UAE: Association for Computational Linguistics.

\bibitem[{Lei et~al.(2025)Lei, Chen, Ye, Cao, Shin, SU, SUO, Gao, Hu, Yin, Zhong, Xiong, Sun, Liu, Wang, and Yu}]{spider2.0}
Lei, F.; Chen, J.; Ye, Y.; Cao, R.; Shin, D.; SU, H.; SUO, Z.; Gao, H.; Hu, W.; Yin, P.; Zhong, V.; Xiong, C.; Sun, R.; Liu, Q.; Wang, S.; and Yu, T. 2025.
\newblock Spider 2.0: Evaluating Language Models on Real-World Enterprise Text-to-{SQL} Workflows.
\newblock In \emph{The Thirteenth International Conference on Learning Representations}.

\bibitem[{Li et~al.(2024{\natexlab{a}})Li, Luo, Chai, Li, and Tang}]{DAWN}
Li, B.; Luo, Y.; Chai, C.; Li, G.; and Tang, N. 2024{\natexlab{a}}.
\newblock The Dawn of Natural Language to SQL: Are We Fully Ready?
\newblock \emph{Proc. VLDB Endow.}, 17(11): 3318–3331.

\bibitem[{Li et~al.(2023{\natexlab{a}})Li, Zhang, Li, and Chen}]{resdsql}
Li, H.; Zhang, J.; Li, C.; and Chen, H. 2023{\natexlab{a}}.
\newblock RESDSQL: decoupling schema linking and skeleton parsing for text-to-SQL.
\newblock In \emph{Proceedings of the Thirty-Seventh AAAI Conference on Artificial Intelligence and Thirty-Fifth Conference on Innovative Applications of Artificial Intelligence and Thirteenth Symposium on Educational Advances in Artificial Intelligence}, AAAI'23/IAAI'23/EAAI'23. AAAI Press.
\newblock ISBN 978-1-57735-880-0.

\bibitem[{Li et~al.(2024{\natexlab{b}})Li, Zhang, Liu, Fan, Zhang, Zhu, Wei, Pan, Li, and Chen}]{CODES}
Li, H.; Zhang, J.; Liu, H.; Fan, J.; Zhang, X.; Zhu, J.; Wei, R.; Pan, H.; Li, C.; and Chen, H. 2024{\natexlab{b}}.
\newblock CodeS: Towards Building Open-source Language Models for Text-to-SQL.
\newblock \emph{Proc. ACM Manag. Data}, 2(3).

\bibitem[{Li et~al.(2023{\natexlab{b}})Li, Hui, Cheng, Qin, Ma, Huo, Huang, Du, Si, and Li}]{li2023graphix}
Li, J.; Hui, B.; Cheng, R.; Qin, B.; Ma, C.; Huo, N.; Huang, F.; Du, W.; Si, L.; and Li, Y. 2023{\natexlab{b}}.
\newblock Graphix-T5: mixing pre-trained transformers with graph-aware layers for text-to-SQL parsing.
\newblock In \emph{Proceedings of the Thirty-Seventh AAAI Conference on Artificial Intelligence and Thirty-Fifth Conference on Innovative Applications of Artificial Intelligence and Thirteenth Symposium on Educational Advances in Artificial Intelligence}, AAAI'23/IAAI'23/EAAI'23. AAAI Press.
\newblock ISBN 978-1-57735-880-0.

\bibitem[{Li et~al.(2023{\natexlab{c}})Li, Hui, Qu, Yang, Li, Li, Wang, Qin, Geng, Huo, Zhou, Ma, Li, Chang, Huang, Cheng, and Li}]{BIRD}
Li, J.; Hui, B.; Qu, G.; Yang, J.; Li, B.; Li, B.; Wang, B.; Qin, B.; Geng, R.; Huo, N.; Zhou, X.; Ma, C.; Li, G.; Chang, K.~C.; Huang, F.; Cheng, R.; and Li, Y. 2023{\natexlab{c}}.
\newblock Can LLM already serve as a database interface? a big bench for large-scale database grounded text-to-SQLs.
\newblock In \emph{Proceedings of the 37th International Conference on Neural Information Processing Systems}, NIPS '23. Red Hook, NY, USA: Curran Associates Inc.

\bibitem[{Liu et~al.(2022)Liu, Hu, Lin, and Wen}]{semantic}
Liu, A.; Hu, X.; Lin, L.; and Wen, L. 2022.
\newblock Semantic Enhanced Text-to-SQL Parsing via Iteratively Learning Schema Linking Graph.
\newblock In \emph{Proceedings of the 28th ACM SIGKDD Conference on Knowledge Discovery and Data Mining}, KDD '22, 1021–1030. New York, NY, USA: Association for Computing Machinery.
\newblock ISBN 9781450393850.

\bibitem[{Liu et~al.(2025)Liu, Tan, Zhong, Xie, Zhao, Wang, Hu, and Li}]{liu2024solid}
Liu, G.; Tan, Y.; Zhong, R.; Xie, Y.; Zhao, L.; Wang, Q.; Hu, B.; and Li, Z. 2025.
\newblock Solid-{SQL}: Enhanced Schema-linking based In-context Learning for Robust Text-to-{SQL}.
\newblock In Rambow, O.; Wanner, L.; Apidianaki, M.; Al-Khalifa, H.; Eugenio, B.~D.; and Schockaert, S., eds., \emph{Proceedings of the 31st International Conference on Computational Linguistics}, 9793--9803. Abu Dhabi, UAE: Association for Computational Linguistics.

\bibitem[{Liu et~al.(2004)Liu, Moore, Yang, and Gray}]{liu2004investigation}
Liu, T.; Moore, A.; Yang, K.; and Gray, A. 2004.
\newblock An Investigation of Practical Approximate Nearest Neighbor Algorithms.
\newblock In Saul, L.; Weiss, Y.; and Bottou, L., eds., \emph{Advances in Neural Information Processing Systems}, volume~17. MIT Press.

\bibitem[{Ma et~al.(2025)Ma, Zhuang, Xu, Jiang, Chen, and Guo}]{ma2025sql}
Ma, P.; Zhuang, X.; Xu, C.; Jiang, X.; Chen, R.; and Guo, J. 2025.
\newblock SQL-R1: Training Natural Language to SQL Reasoning Model By Reinforcement Learning.
\newblock arXiv:2504.08600.

\bibitem[{Maamari et~al.(2024)Maamari, Abubaker, Jaroslawicz, and Mhedhbi}]{maamari2024death}
Maamari, K.; Abubaker, F.; Jaroslawicz, D.; and Mhedhbi, A. 2024.
\newblock The Death of Schema Linking? Text-to-{SQL} in the Age of Well-Reasoned Language Models.
\newblock In \emph{NeurIPS 2024 Third Table Representation Learning Workshop}.

\bibitem[{Pourreza et~al.(2025{\natexlab{a}})Pourreza, Li, Sun, Chung, Talaei, Kakkar, Gan, Saberi, Ozcan, and Arik}]{chase}
Pourreza, M.; Li, H.; Sun, R.; Chung, Y.; Talaei, S.; Kakkar, G.~T.; Gan, Y.; Saberi, A.; Ozcan, F.; and Arik, S. 2025{\natexlab{a}}.
\newblock CHASE-SQL: Multi-Path Reasoning and Preference Optimized Candidate Selection in Text-to-SQL.
\newblock In Yue, Y.; Garg, A.; Peng, N.; Sha, F.; and Yu, R., eds., \emph{International Conference on Representation Learning}, volume 2025, 60385--60415.

\bibitem[{Pourreza and Rafiei(2023)}]{DIN-SQL}
Pourreza, M.; and Rafiei, D. 2023.
\newblock DIN-SQL: decomposed in-context learning of text-to-SQL with self-correction.
\newblock In \emph{Proceedings of the 37th International Conference on Neural Information Processing Systems}, NIPS '23. Red Hook, NY, USA: Curran Associates Inc.

\bibitem[{Pourreza et~al.(2025{\natexlab{b}})Pourreza, Talaei, Sun, Wan, Li, Mirhoseini, Saberi, and Arik}]{pourreza2025reasoning}
Pourreza, M.; Talaei, S.; Sun, R.; Wan, X.; Li, H.; Mirhoseini, A.; Saberi, A.; and Arik, S.~O. 2025{\natexlab{b}}.
\newblock Reasoning-SQL: Reinforcement Learning with SQL Tailored Partial Rewards for Reasoning-Enhanced Text-to-SQL.
\newblock arXiv:2503.23157.

\bibitem[{Qu et~al.(2024)Qu, Li, Li, Qin, Huo, Ma, and Cheng}]{TA-SQL}
Qu, G.; Li, J.; Li, B.; Qin, B.; Huo, N.; Ma, C.; and Cheng, R. 2024.
\newblock Before Generation, Align it! A Novel and Effective Strategy for Mitigating Hallucinations in Text-to-{SQL} Generation.
\newblock In Ku, L.-W.; Martins, A.; and Srikumar, V., eds., \emph{Findings of the Association for Computational Linguistics ACL 2024}, 5456--5471. Bangkok, Thailand and virtual meeting: Association for Computational Linguistics.

\bibitem[{Safdarian et~al.(2025)Safdarian, Mohammadi, Jahanbakhsh, Naderi, and Faili}]{safdarian2025schemagraphsql}
Safdarian, A.; Mohammadi, M.; Jahanbakhsh, E.; Naderi, M.~S.; and Faili, H. 2025.
\newblock SchemaGraphSQL: Efficient Schema Linking with Pathfinding Graph Algorithms for Text-to-SQL on Large-Scale Databases.
\newblock arXiv:2505.18363.

\bibitem[{Shi et~al.(2025)Shi, Tang, Zhang, Zhang, and Yang}]{shi2024survey}
Shi, L.; Tang, Z.; Zhang, N.; Zhang, X.; and Yang, Z. 2025.
\newblock A Survey on Employing Large Language Models for Text-to-SQL Tasks.
\newblock \emph{ACM Comput. Surv.}, 58(2).

\bibitem[{Talaei et~al.(2024)Talaei, Pourreza, Chang, Mirhoseini, and Saberi}]{chess}
Talaei, S.; Pourreza, M.; Chang, Y.-C.; Mirhoseini, A.; and Saberi, A. 2024.
\newblock CHESS: Contextual Harnessing for Efficient SQL Synthesis.
\newblock arXiv:2405.16755.

\bibitem[{Wang et~al.(2025)Wang, Ren, Yang, Liang, Bai, Chai, Yan, Zhang, Yin, Sun, and Li}]{MAC-SQL}
Wang, B.; Ren, C.; Yang, J.; Liang, X.; Bai, J.; Chai, L.; Yan, Z.; Zhang, Q.-W.; Yin, D.; Sun, X.; and Li, Z. 2025.
\newblock {MAC}-{SQL}: A Multi-Agent Collaborative Framework for Text-to-{SQL}.
\newblock In Rambow, O.; Wanner, L.; Apidianaki, M.; Al-Khalifa, H.; Eugenio, B.~D.; and Schockaert, S., eds., \emph{Proceedings of the 31st International Conference on Computational Linguistics}, 540--557. Abu Dhabi, UAE: Association for Computational Linguistics.

\bibitem[{Wang et~al.(2020)Wang, Shin, Liu, Polozov, and Richardson}]{RAT-SQL}
Wang, B.; Shin, R.; Liu, X.; Polozov, O.; and Richardson, M. 2020.
\newblock {RAT-SQL}: Relation-Aware Schema Encoding and Linking for Text-to-{SQL} Parsers.
\newblock In Jurafsky, D.; Chai, J.; Schluter, N.; and Tetreault, J., eds., \emph{Proceedings of the 58th Annual Meeting of the Association for Computational Linguistics}, 7567--7578. Online: Association for Computational Linguistics.

\bibitem[{Wang et~al.(2023)Wang, Wei, Schuurmans, Le, Chi, Narang, Chowdhery, and Zhou}]{wang2022self}
Wang, X.; Wei, J.; Schuurmans, D.; Le, Q.~V.; Chi, E.~H.; Narang, S.; Chowdhery, A.; and Zhou, D. 2023.
\newblock Self-Consistency Improves Chain of Thought Reasoning in Language Models.
\newblock In \emph{The Eleventh International Conference on Learning Representations}.

\bibitem[{Wang, Liu, and Yang(2025)}]{LINKALIGN}
Wang, Y.; Liu, P.; and Yang, X. 2025.
\newblock {L}ink{A}lign: Scalable Schema Linking for Real-World Large-Scale Multi-Database Text-to-{SQL}.
\newblock In \emph{Proceedings of the 2025 Conference on Empirical Methods in Natural Language Processing}, 977--991. Suzhou, China: Association for Computational Linguistics.
\newblock ISBN 979-8-89176-332-6.

\bibitem[{Yang et~al.(2024)Yang, Su, Li, Li, Mao, Liu, and Zhao}]{SQL-TO-SCHEMA}
Yang, S.; Su, Q.; Li, Z.; Li, Z.; Mao, H.; Liu, C.; and Zhao, R. 2024.
\newblock SQL-to-Schema Enhances Schema Linking in Text-to-SQL.
\newblock In Strauss, C.; Amagasa, T.; Manco, G.; Kotsis, G.; Tjoa, A.~M.; and Khalil, I., eds., \emph{Database and Expert Systems Applications}, 139--145. Cham: Springer Nature Switzerland.
\newblock ISBN 978-3-031-68309-1.

\bibitem[{Yu et~al.(2018)Yu, Zhang, Yang, Yasunaga, Wang, Li, Ma, Li, Yao, Roman, Zhang, and Radev}]{spider1.0}
Yu, T.; Zhang, R.; Yang, K.; Yasunaga, M.; Wang, D.; Li, Z.; Ma, J.; Li, I.; Yao, Q.; Roman, S.; Zhang, Z.; and Radev, D. 2018.
\newblock {S}pider: A Large-Scale Human-Labeled Dataset for Complex and Cross-Domain Semantic Parsing and Text-to-{SQL} Task.
\newblock In Riloff, E.; Chiang, D.; Hockenmaier, J.; and Tsujii, J., eds., \emph{Proceedings of the 2018 Conference on Empirical Methods in Natural Language Processing}, 3911--3921. Brussels, Belgium: Association for Computational Linguistics.

\end{thebibliography}

\clearpage
\appendix
\section{Details of Environment Construction} \label{Details of Environment Construction}

\begin{algorithm}
\caption{AutoLink: Autonomous Schema Exploration and Expansion}
\label{alg:autolink}
\begin{algorithmic}[1]
\REQUIRE Database Environment $\mathcal{E}_{\text{DB}}$, Schema Vector Store Environment $\mathcal{E}_{\text{VS}}$
\REQUIRE Instruction Prompt $I$, User Question $Q$, Full Database Schema $S_{\text{full}}$, All Table Names $T$ extracted from $S_{\text{full}}$
\REQUIRE Maximum Interaction Turns $T_{\text{max}}$, Initial Retrieval Count $n$, Targeted Retrieval Count $m$
\ENSURE Final Linked Schema $S_{\text{linked}}$
\STATE \textbf{Initialize:}
\STATE \quad History $H \leftarrow \text{empty string}$
\STATE \quad Current Linked Schema $S_{\text{linked}} \leftarrow \emptyset$
\STATE \quad Excluded Columns $\mathcal{C}_{\text{excl}} \leftarrow \emptyset$
\STATE \textbf{Step 1: Initial Schema Generation}
\STATE \quad $S_{\text{initial}} \leftarrow \mathcal{E}_{\text{VS}}(Q, n, \mathcal{C}_{\text{excl}})$ \COMMENT{Retrieve top-$n$ columns from }
\STATE \quad $S_{\text{linked}} \leftarrow S_{\text{initial}}$
\STATE \quad $\mathcal{C}_{\text{excl}} \leftarrow \text{columns in } S_{\text{initial}}$
\STATE \quad $H_0 \leftarrow \text{Construct initial context with } I, Q, T, S_{\text{initial}}$
\STATE \textbf{Step 2: Agent-Environment Interaction}
\FOR{$t = 1$ \TO $T_{\text{max}}$}
    \STATE \quad Construct current prompt $P_t \leftarrow H_{t-1}$
    \STATE \quad $(\theta_t, A_t) \leftarrow \pi(P_t)$ \COMMENT{LLM agent generates reasoning and actions}
    \STATE \quad $O_t \leftarrow \text{empty string}$ \COMMENT{Initialize observations for current turn}
    \STATE \quad $\text{stop\_flag} \leftarrow \text{FALSE}$
    \FOR{each action $a \in A_t$}
        \IF{$a = \text{@explore\_schema}(\text{sql\_query})$}
            \STATE \quad $R_{\text{SQL}} \leftarrow \mathcal{E}_{\text{DB}}(\text{sql\_query})$
            \STATE \quad Append $R_{\text{SQL}}$ to $O_t$
        \ELSIF{$a = \text{@retrieve\_schema}(\text{nl\_query})$}
            \STATE \quad $S_{\text{retrieved}} \leftarrow \mathcal{E}_{\text{VS}}(\text{nl\_query}, m, \mathcal{C}_{\text{excl}})$
            \STATE \quad Append $S_{\text{retrieved}}$ to $O_t$
        \ELSIF{$a = \text{@verify\_schema}(\text{sql\_query})$}
            \STATE \quad $R_{\text{SQL}} \leftarrow \mathcal{E}_{\text{DB}}(\text{sql\_query})$
            \STATE \quad Append $R_{\text{SQL}}$ to $O_t$
        \ELSIF{$a = \text{@add\_schema}(\text{schemas})$}
            \STATE \quad $S_{\text{added}} \leftarrow \text{extract full metadata of schemas}$
            \STATE \quad $S_{\text{linked}} \leftarrow S_{\text{linked}} \cup S_{\text{added}}$
            \STATE \quad $\mathcal{C}_{\text{excl}} \leftarrow \mathcal{C}_{\text{excl}} \cup \text{columns in } S_{\text{added}}$
        \ELSIF{$a = \text{@stop\_action}()$}
            \STATE \quad $\text{stop\_flag} \leftarrow \text{TRUE}$
        \ENDIF
    \ENDFOR
    \STATE \quad $H_t \leftarrow H_{t-1} \cup \{(\theta_t, A_t, O_t)\}$ \COMMENT{Update history}
    \STATE \quad Update prompt $P_{t+1}$ with $H_t$
    \IF{$\text{stop\_flag}$ is TRUE}
        \STATE \quad \textbf{BREAK}
    \ENDIF
\ENDFOR
\RETURN $S_{\text{linked}}$
\STATE \textbf{Step 3: SQL Generation}
\STATE \quad $SQL_{\text{predicted}} \leftarrow \pi_{\text{SQL}}(Q, S_{\text{linked}})$ \COMMENT{The entire SQL generation process operates \textbf{only on} the final simplified linked schema $S_{\text{linked}}$.}
\end{algorithmic}
\end{algorithm}

This section provides a detailed exposition of the construction and operational principles behind the two core external environments within our \textbf{AutoLink} framework: the \textbf{\emph{Database Environment ($\mathcal{E}_{\text{DB}}$)}} and the \textbf{\emph{Schema Vector Store Environment ($\mathcal{E}_{\text{VS}}$)}}. These environments are designed to furnish the autonomous agent with the necessary infrastructure for iterative exploration, schema verification, and efficient semantic retrieval. 

\paragraph{Database Environment ($\mathcal{E}_{\text{DB}}$)} ~ {The output $R_{\text{SQL}}$ from $\mathcal{E}_{\text{DB}}$, generated for any arbitrary SQL query, is meticulously structured to offer clear and actionable feedback to the agent. This environment is designed for exploratory and experimental SQL queries, thus discouraging overly complex or long-running operations. To ensure efficient interaction and manage the LLM's context length, strict constraints are enforced: 1) The data payload in $R_{\text{SQL}}$ is truncated to a maximum of \textbf{5 rows}, and 2) Each SQL query's execution is capped at \textbf{120 seconds}, with automatic termination if this limit is exceeded. The structure of $R_{\text{SQL}}$ adapts to the query's execution outcome, providing detailed feedback for successful operations (non-empty or empty result set) and clear error messages for failures or timeouts. Specific examples are shown below. 

\begin{itemize}

    \item \textbf{Successful Query Execution: Non-Empty Result Set} \quad When a query executes successfully and returns results within the time limit, $R_{\text{SQL}}$ provides information structured as follows: 
    
\begin{minted}[
frame=lines,
framesep=2mm,
linenos,
gobble=0,
fontsize=\small,
bgcolor=lightgray!10,
breaklines=true
]{markdown}
[Total rows: 123, Execution time: 0.05s, Top-5 rows are shown bellow]
Column1 | Column2
--------|--------
Value1  | ValueA
Value2  | ValueB
Value3  | ValueC
Value4  | ValueD
Value5  | ValueE
118 rows truncated ...
\end{minted}

     \item \textbf{Successful Query Execution: Empty Result Set} \quad If a query executes successfully within the time limit but yields no rows (e.g., a SELECT query with no matches), $R_{\text{SQL}}$ provides a distinct message:

\begin{minted}[
    frame=lines, % 添加边框
    framesep=2mm, % 边框与代码的距离
    linenos, % 显示行号
    gobble=0, % 移除行首空格
    fontsize=\small, % 字体大小
    bgcolor=lightgray!10, % 背景颜色
    breaklines=true % 自动换行
]{markdown}
[No data found for the specified query, Execution time: 0.2s]
\end{minted}

     \item \textbf{SQL Execution Timeout} \quad If a query exceeds the execution time limit, $R_{\text{SQL}}$ returns a specific error message indicating the timeout:

\begin{minted}[
    frame=lines, % 添加边框
    framesep=2mm, % 边框与代码的距离
    linenos, % 显示行号
    gobble=0, % 移除行首空格
    fontsize=\small, % 字体大小
    bgcolor=lightgray!10, % 背景颜色
    breaklines=true % 自动换行
]{markdown}
[[ERROR: SQL execution timed out after 120 seconds]]
\end{minted}

     \item \textbf{SQL Execution Error} For any other SQL execution error (e.g., \emph{syntax error}, \emph{non-existent column}), $R_{\text{SQL}}$ contains the specific, verbatim error message generated by the SQL execution engine. This direct feedback is crucial for the agent to debug its queries and refine its understanding of the schema:
\begin{minted}[
    frame=lines, % 添加边框
    framesep=2mm, % 边框与代码的距离
    linenos, % 显示行号
    gobble=0, % 移除行首空格
    fontsize=\small, % 字体大小
    bgcolor=lightgray!10, % 背景颜色
    breaklines=true % 自动换行
]{markdown}
[ERROR: column "user_id" does not exist]
\end{minted}

\end{itemize}

}

\paragraph{Schema Vector Store Environment ($\mathcal{E}_{\text{VS}}$)}\label{appendix:vector-store-env} ~ {To facilitate semantic retrieval over schema elements, we construct a vector database indexing all columns in the schema. In practice, special care is taken to handle partitioned tables prevalent in BigQuery and similar data warehouses. The overall setup procedure consists of the following steps:

\begin{itemize}
    \item \textbf{Handling of Partitioned Tables} \quad In data warehouses like BigQuery, it is common to encounter partitioned tables (e.g., \emph{`table\_20230101`}, \emph{`table\_20230102`}, etc.) that share an identical underlying column structure but differ only in their names, typically reflecting a time-based partition. To maximize efficiency and avoid redundant embeddings for such tables, we implement a merging strategy. We first identify and group all tables that possess \textbf{exactly the same column schemas}. For a group of such partitioned tables $\{ t_{p,1}, t_{p,2},... \}$ that effectively represent the same logical entity with schema $\{ c_1, ..., c_n \}$, each unique column $c_i$ from this shared schema is processed and \textbf{embedded only once}. This merging strategy significantly reduces the number of redundant entries in the vector store while preserving all necessary table context for retrieval. For example, in the \textbf{Spider 2.0-Lite}~\cite{spider2.0} benchmark, after applying this merging strategy, each database contains an \textbf{average of 559 columns, with a maximum of 6161 columns}. Crucially, without merging partitioned tables, the total number of columns could be as high as \textbf{100,000}.
    
    \item \textbf{Textual Representation for each Column} \quad For each column $c_i \in S_{\text{full}}$ (after the merging process for partitioned tables), we construct a textual document by concatenating its core metadata. For each column, the constructed document includes: 1) \textbf{\emph{column name}}; 2)  \textbf{\emph{associated table names list}}, i.e., a list containing the names of all tables that share this exact column schema (e.g., all table names for partitioned tables that were grouped); 3)  \textbf{\emph{column data type}}; 4)  \textbf{\emph{column description}} (if available) and 5) \textbf{\emph{declaration of primary and foreign keys}}. An example column textual document is shown below:
    
\begin{minted}[
frame=lines, % 添加边框
framesep=2mm, % 边框与代码的距离
linenos, % 显示行号
gobble=0, % 移除行首空格
fontsize=\small, % 字体大小
bgcolor=lightgray!10, % 背景颜色
breaklines=true % 自动换行
]{markdown}
Column: visitNumber;
Table: [ga_sessions_20170720, ga_sessions_20170521, ...];
Type: INT64;
Description: The session number for this user;
Primary Key: None;
Foreign Key: None;
\end{minted}

    \item \textbf{Column Vector Store Indexing} \quad Each column document (with all texts are lowercased) is embedded into a 1024-dimensional dense vector using the bge-large-en-v1.5~\cite{chen2024bge} model, a BERT-based~\cite{devlin2019bert} open-source embedding model. These column embeddings are then stored in a vector database, specifically using Faiss~\cite{johnson2019billion} for its Approximate Nearest Neighbor (ANN) \cite{liu2004investigation} index. Each vector entry maintains a link to its comprehensive column metadata for downstream retrieval and formatting. To \textbf{quantify the construction overhead of this schema vector store environment per database}, we measure the time required for its construction on \textbf{one H100 GPU} using a \textbf{batch size of 1024}. For the \textbf{Spider 2.0-Lite}~\cite{spider2.0} benchmark, encompassing \textbf{158 databases}, the total time required for programmatic column document construction, embedding, and indexing into the vector database are \textbf{222.8} seconds, averaging a mere \textbf{1.4 seconds per database}. Similarly, for the \textbf{BIRD}~\cite{BIRD} benchmark, comprising \textbf{11 databases}, the complete environment setup takes \textbf{4.4} seconds, averaging an even faster \textbf{0.4 seconds per database}. It underscores the rapid and low-cost deployment capabilities of our schema vector store. 
    
    \item  \textbf{Column Retrieval} \quad Upon constructing the column vector store, a given natural language query is first encoded into a query vector using the same text encoder (i.e., bge-large-en-v1.5). This query vector is then utilized to retrieve the top-$K$ most semantically similar columns from the vector store via Approximate Nearest Neighbor (ANN) search, ranked by \textbf{cosine similarity}, with a built-in mechanism to exclude any columns that have been previously returned. \textbf{On average, encoding the query and performing the retrieval takes only 0.05 seconds on one H800 GPU}. For these retrieved columns, we identify their corresponding tables, or all associated table names for partitioned tables, and form a relevant schema subset, which is formatted and presented in the M-Schema~\cite{gao2024preview} style. \textbf{M-Schema} is a semi-structured textual representation for schema, explicitly identifies hierarchical relationships between databases, tables, and columns using specific tokens, and provides comprehensive details such as column names, data types, primary/foreign keys, detailed descriptions, and sampled values. Bellow is an example of M-Schema for a partitioned table:

\begin{minted}[
frame=lines, % 添加边框
framesep=2mm, % 边框与代码的距离
linenos, % 显示行号
gobble=0, % 移除行首空格
fontsize=\small, % 字体大小
bgcolor=lightgray!10, % 背景颜色
breaklines=true % 自动换行
]{markdown}
[DB_ID] ga360
# Table ga_sessions_20160810
(visitNumber: INT64, Examples:[2,1,1,1,1], The session number for this user. If this is the first session, then this is set to 1)
(fullVisitorId: STRING, Examples:["1906","7880","5836",
"6743","0244"], The unique visitor ID.)
(date: STRING, Examples:["20160810","20160810",
"20160810","20160810","20160810"]: The date of the session in YYYYMMDD format.)
(visitStartTime: INT64, Examples:["1470817379","1470856696",
"1470873918","1470816856","1470883270"], The timestamp (expressed as POSIX time).)
\end{minted}

\end{itemize}



\subsection{Details of Action Space}

In the main paper, we introduce the core actions that empower \textbf{AutoLink}'s autonomous agent. This section provides a more comprehensive overview of three actions (i.e., \textbf{@explore\_schema}, \textbf{@retrieve\_schema} and \textbf{@add\_schema}) within the agent's action space. For each action, we offer concrete examples of its usage and detail its specific utility. Readers can also refer to AutoLink's pseudocode in Algorithm~\ref{alg:autolink} for a better understanding of these actions.

\noindent\textbf{1. @explore\_schema} \quad This action is designed to interact with the Database Environment ($\mathcal{E}_{\text{DB}}$) by executing exploratory SQL queries mainly against the full schema metadata $S_{\text{full}}$. Exploratory queries under this action can be categorized into two main aspects: exploring value distributions and exploring schema structure. For example, in terms of exploring value distributions, the agent can execute queries to verify specific values, sample data, or analyze value ranges, which helps confirm whether the values involved in the problem exist in the target columns:
\begin{minted}[
frame=lines,
framesep=2mm,
linenos,
gobble=0,
fontsize=\small,
bgcolor=lightgray!10,
breaklines=true
]{sql}
-- Check sample values containing specific keywords (e.g., "INTOX")
SELECT DISTINCT descript
FROM incidents_2016
WHERE descript LIKE '%INTOX%'
LIMIT 5;

-- Sample values to understand data format
SELECT date, timestamp, time
FROM incidents_2016
LIMIT 5;

-- Analyze value range of a date column
SELECT MIN(date), MAX(date)
FROM incidents_2016;
\end{minted}

In terms of exploring schema structure, the agent can check table structures, search for relevant tables or columns using metadata queries, which helps clarify the database's organizational structure. Note that \textbf{INFORMATION\_SCHEMA} is a standard metadata repository supported by most relational databases (e.g.,BigQuery, Snowflake) for querying schema details, while \textbf{PRAGMA} is a database-specific command primarily used in SQLite for retrieving metadata like table structures. Specific exploration examples are as follows:
\begin{minted}[
frame=lines,
framesep=2mm,
linenos,
gobble=0,
fontsize=\small,
bgcolor=lightgray!10,
breaklines=true
]{sql}
-- Explore table structure by sampling rows to observe column types and data patterns
SELECT * FROM inpatient_charges_2014
LIMIT 5;

-- Search for columns related to target semantics using INFORMATION_SCHEMA
SELECT column_name
FROM new_york.INFORMATION_SCHEMA.COLUMNS
WHERE column_name LIKE '%trips%';

-- Filter columns by dual conditions (table attributes + column semantics) via metadata
SELECT column_name
FROM census.INFORMATION_SCHEMA.COLUMNS
WHERE table_name LIKE '%tract%' AND table_name LIKE '%2018%'
AND LOWER(column_name) LIKE '%income%'
LIMIT 5;

-- Identify relevant reference tables by keyword matching in table names
FROM new_york_taxi.INFORMATION_SCHEMA.TABLES
WHERE LOWER(table_name) LIKE '%zone%' OR LOWER(table_name) LIKE '%borough%'
LIMIT 5;

-- check column details of a table using database-specific metadata commands (e.g., PRAGMA for SQLite)
PRAGMA table_info(results);
\end{minted}
\noindent\textbf{2. @retrieve\_schema} \quad The action empowers the agent to actively search for missing schema elements using the Schema Vector Store Environment ($\mathcal{E}_{\text{VS}}$). Unlike simple query rewrite methods such as CHESS~\cite{chess}, this action is dynamic: the agent formulates a new natural language query for the vector store by combining the original user question with contextual information, including all table names and partially linked schema elements. Crucially, this query can directly be an \textbf{\emph{inferred virtual column name}}, \textbf{\emph{a description of a potentially missing column}}, or an \textbf{\emph{abstract concept/phrase}} that the agent generates to target the discovery of missing schema elements. This capability is vital because it transcends the limitations of simple query rewriting, which often struggles with ambiguous terms or when relevant schema elements are not directly discoverable from the literal words in the user's question, enabling a more sophisticated and targeted search. 

For example, consider a user asking, \emph{``What's the score?"} On its own, this question is ambiguous; \emph{``score"} could refer to a \emph{game score}, a \emph{credit score}, a \emph{performance score}, or a \emph{test score}. A simple query rewriter would struggle to map \emph{``score"} to a specific, actionable database column, as its meaning is entirely context-dependent. However, if the agent knows there is a table named \textbf{\emph{students}}, the question's intent immediately becomes clear: the user is asking about \emph{student academic scores}. In this context, the semantic search can specifically target columns like \emph{exam\_score}, \emph{quiz\_score}, or \emph{final\_grade}, even if these specific columns are not explicitly present in the partial schema. 


\begin{minted}[
    frame=lines, % 添加边框
    framesep=2mm, % 边框与代码的距离
    linenos, % 显示行号
    gobble=0, % 移除行首空格
    fontsize=\small, % 字体大小
    bgcolor=lightgray!10, % 背景颜色
    breaklines=true % 自动换行
]{sql}
@retrieve_schema(`exam score, quiz score or final grade`);
\end{minted}

\noindent\textbf{3. @verify\_schema} \quad This action takes the currently linked schema elements $S_{\text{linked}}$ as a working hypothesis and constructs a \textbf{minimal executable} SQL query to test whether this hypothesis is sufficient to answer the user's question. The purpose of this action is not to obtain the final query result, but to convert the database engine's execution outcome—success or error messages—into high-precision diagnostic signals. For example, \emph{no such column: X} clearly indicates a missing column X; \emph{no such table: T} points to an unlinked table. Therefore, @verify\_schema often forms a feedback loop with @retrieve\_schema and @explore\_schema: errors provide semantic clues for targeted retrieval, which are then incorporated into $S_{\text{linked}}$ via @add\_schema, followed by re-verification.

An example of verification is as follows, suppose the user asks, \emph{``How many arrests occurred in 2016?”} The agent hypothesizes that the relevant table is \emph{incidents\_2016} and that the arrest indicator column might be \emph{is\_arrest}. It then issues the following verification query:
\begin{minted}[
    frame=lines, % 添加边框
    framesep=2mm, % 边框与代码的距离
    linenos, % 显示行号
    gobble=0, % 移除行首空格
    fontsize=\small, % 字体大小
    bgcolor=lightgray!10, % 背景颜色
    breaklines=true % 自动换行
]{sql}
-- Verification query (minimal hypothesis)
SELECT COUNT(*)
FROM incidents_2016
WHERE is_arrest = 1;
\end{minted}
If the execution returns \emph{Error: no such column: is\_arrest}, the agent treats this as a precise signal and invokes @retrieve\_schema (e.g., \emph{searching for a ``column indicating arrest status”}). After adding the retrieved column via @add\_schema, it re-runs a simplified verification query to confirm that the schema has been sufficiently corrected.

\noindent\textbf{4. @add\_schema} \quad This is the agent's mechanism for committing discoveries. After actions with feedback yields new, relevant schema elements, the agent can adopt this action to add them to its final candidate set. We encourage agent to use this action frequently to add sufficient schema elements to improve schema linking recall, using the format of \emph{table\_name.column\_name}, as shown below. 

\begin{minted}[
    frame=lines, % 添加边框
    framesep=2mm, % 边框与代码的距离
    linenos, % 显示行号
    gobble=0, % 移除行首空格
    fontsize=\small, % 字体大小
    bgcolor=lightgray!10, % 背景颜色
    breaklines=true % 自动换行
]{sql}
@add_schema(`orders.id; orders.prod_sku; users.is_active; products.cat_id`);
\end{minted}

\section{Detail of SQL Generation}
\label{appendix:sql-generation}

Given the final linked schema \( S_{\text{linked}} \) produced by our agent-based retrieval and linking process, we formalize the subsequent SQL generation step as a conditional sequence generation problem. The overall pipeline consists of multiple modules designed to enhance both syntactic and semantic correctness:

\paragraph{SQL Candidate Generation via LLM Policy}~{Let \(\pi_{\text{SQL}}\) denote a large language model (LLM) policy. Given the user question \( Q \) and the contextualized schema \( S_{\text{linked}} \), we generate a set of \( N \) SQL candidates (5 in this paper):
\begin{equation}
\{\texttt{SQL}_i\}_{i=1}^N = \pi_{\text{SQL}}(Q, S_{\text{linked}})
\end{equation}
where \( N \) is the number of self-consistency samples~\cite{wang2022self}. Each candidate is independently generated by sampling from the LLM in stochastic decoding mode (temperature sampling).}

\paragraph{Iterative Syntactic Correction} ~ {Each sampled SQL candidate \(\texttt{SQL}_i\) is refined through an iterative syntactic correction process, also facilitated by the LLM policy \(\pi_{\text{SQL}}\) in a multi-turn dialogue. We denote by \(t\) the maximum number of dialogue turns. Let \(\mathcal{C}_0\) be the initial conversation context, which includes the user query \(Q\), the linked schema \(S_{\text{linked}}\), and the initial SQL candidate \(\texttt{SQL}_i^{(0)} = \texttt{SQL}_i\). At each turn \(j\), we update the conversation context \(\mathcal{C}_j\) by incorporating the previously proposed SQL statement and its execution error:
\begin{equation}
\mathcal{C}_j = \mathcal{C}_{j-1} \,\cup\, \{(\texttt{SQL}_i^{(j-1)}, \text{error}_j)\}.
\end{equation}
The LLM then produces the revised SQL statement:
\begin{equation}
\texttt{SQL}_i^{(j)} = \pi_{\text{SQL}}(\mathcal{C}_j).
\end{equation}
This process continues until \(\texttt{SQL}_i^{(j)}\) successfully executes without errors or until the maximum number of dialogue turns \(t\) is reached. The final corrected version after \(k\) turns is denoted by:
\begin{equation}
\texttt{SQL}_i^{\text{corr}} = \texttt{SQL}_i^{(k)},
\end{equation}
where \(k \le t\) represents the iteration at which the statement is deemed correct or the dialogue terminates.}

\paragraph{Majority Voting via Execution-Based Output Grouping} ~ {To further enhance robustness, we employ a majority voting strategy grounded in the execution results of SQL candidates. The complete procedure is as follows:

\begin{itemize}
    \item \textbf{Step 1: Execution-Based Grouping.}
    Consider the set of syntactically valid SQL candidates $\mathcal{F} = \{ \texttt{SQL}_i^{\text{corr}} \}_{i=1}^N$. For each candidate, we execute it on the database and group all candidates that yield identical outputs (i.e., same result set). Formally, let $\mathcal{G}_j$ denote the set of candidates corresponding to the $j$-th unique query output.

    \item \textbf{Step 2: Group Selection by Majority.}
    Identify the group(s) with the largest cardinality. Let $\mathcal{G}_{\max}$ denote \emph{the set of group(s)} attaining the maximal size:
    \begin{equation}
        \mathcal{G}_{\max} = \left\{ \mathcal{G}_j \mid |\mathcal{G}_j| = \max_k |\mathcal{G}_k| \right\}
    \end{equation}

    \item \textbf{Step 3: Final SQL Selection.}
    \begin{itemize}
        \item \textit{Single Majority Group:} If $|\mathcal{G}_{\max}| = 1$ (i.e., only one group has the majority count), we randomly select one SQL from this group as the final output.
        \item \textit{Multiple Majority Groups (Tie):} If multiple groups share the maximal size, we select one representative SQL candidate from each tied group and aggregate these representatives into the set $\mathcal{S}_{\text{tie}}$. For every unordered pair $(\texttt{SQL}_a, \texttt{SQL}_b)$ in $\mathcal{S}_{\text{tie}}$, we prompt an LLM (provided with the user question, schema, candidate SQL, and their execution results) to select the better SQL between them. Each win counts as one point. The SQL with the highest total number of pairwise wins is selected. In the rare event of a tie in pairwise wins, we randomly select one among the top scorers.
    \end{itemize}
\end{itemize}
}

Formally, the final selection can be expressed as:
\begin{equation}
\texttt{SQL}^{*} =
\begin{cases}
\text{RandomSelect}(\mathcal{G}_{\max}), & \text{if } |\mathcal{G}_{\max}| = 1 \\
\underset{\texttt{SQL} \in \mathcal{S}_{\text{tie}}}{\operatorname{argmax}} \, \text{PairwiseWins}(\texttt{SQL}), & \text{otherwise}
\end{cases}
\end{equation}
where $\text{PairwiseWins}(\texttt{SQL})$ denotes the number of times a SQL candidate is preferred over others in LLM-based pairwise comparisons, and $\mathcal{S}_{\text{tie}}$ is the set of all SQLs in the tied majority groups.

\section{More Experiment Details}
\subsection{Experimental Setup}
In the main experimental results of schema linking, we set top-$n$ to 100 for initial schema retrieval in Spider 2.0-lite, while for the smaller Bird dataset, we set top-$n$ to 30. For the \textbf{@retrieve\_schema} action, top-$m$ is fixed at 3, and the maximum interaction turn limit is set to 10. Additionally, for the DE-SL and CE-SL methods, we set retrieval top-$k$ to 200 on Spider 2.0-lite and 40 on the Bird dataset. The number of sampling decoding is set to 5. 


In the SQL generation phase, we adopt a self-consistency sampling approach with a temperature setting of 1.0, generating \textbf{5} SQL candidates.  Subsequently, these candidates undergo iterative syntactic correction, with a maximum of 5 dialogue turns allowed for each candidate to integrate execution feedback and refine the SQL until it successfully executes. On the Bird dataset, we utilize DeepSeek-V3 for generating SQL. However, on the Spider 2.0-Lite dataset, which demands more complex logical reasoning, we employ DeepSeek-R1.

\subsection{Benchmark}

\begin{itemize}
\item \textbf{Bird}~\cite{BIRD} \quad is a cross-domain dataset designed to evaluate the impact of extensive database contents on text-to-SQL parsing. It comprises over 12,751 unique SQL question pairs, spans 95 large databases, and has a total size of 33.4 GB. The dataset encompasses more than 37 specialized domains. All our experiments on BIRD are conducted on the Bird Dev dataset.

\item \textbf{Spider 2.0-Lite}~\cite{spider2.0} \quad serves as a benchmark to assess the performance of language models on complex enterprise-level text-to-SQL tasks. As an upgrade from Spider 1.0 \cite{spider1.0}, it focuses on more intricate SQL generation tasks across various databases and SQL dialects. Spider 2 includes multiple versions--Spider 2.0, Spider 2.0-Lite, and Spider 2.0-Snow--tailored for different database systems such as BigQuery, Snowflake, and SQLite. Given the inherent complexity of Spider 2.0-Lite, which supports these three distinct database dialects, all our experiments concerning Spider 2.0 were exclusively conducted on the Spider 2.0-Lite dataset. Its main features include a complex environment with over 3,000 columns, multi-step SQL generation requiring handling long contexts and complex reasoning, and a notably challenging nature, where even advanced models like GPT-4 achieve only 6.0\% accuracy, significantly lower than the 86.6\% success rate of Spider 1.0. Our schema linking evaluations are performed on a subset of Spider 2.0-Lite. This subset, comprising 250 examples, is crucial as it provides the necessary ground truth SQL for evaluating schema linking recall.

\end{itemize}

\subsection{Baselines}
\begin{itemize}
\item \textbf{DE-SL:} DE-SL employs a dual-encoder architecture to encode the question and schema elements separately. The relevance between questions and schema components is computed based on the similarity of their encoded representations.
\item \textbf{CE-SL:} CE-SL uses a cross-encoder model that jointly encodes pairs of question tokens and schema elements. This allows the model to directly model intricate interactions between the question and schema components, generally leading to higher linking accuracy at the cost of increased computation.
\item \textbf{MCS-SQL:} MCS-SQL \cite{MCS-SQL} expands the schema linking search space by utilizing multiple prompts and leveraging the sensitivity of large language models (LLMs) to in-context learning (ICL) exemplars. By decoding the LLM multiple times with diverse prompts, MCS-SQL obtains a broader range of candidate schema elements and ultimately selects the most relevant ones for the given question.
\item \textbf{SQL-To-Schema:} SQL-to-Schema \cite{SQL-TO-SCHEMA} generates an initial SQL query by leveraging the complete database schema. It then extracts the involved tables and columns from the generated SQL to construct a concise schema tailored to the input question.
\item \textbf{RSL-SQL:} RSL-SQL \cite{RSL-SQL} integrates several techniques such as bidirectional schema linking, contextual information enhancement, a binary selection strategy, and multi-turn self-correction. These approaches collectively enable more robust schema linking, leading to improved performance on text-to-SQL tasks.
\item \textbf{LinkAlign:} LinkAlign \cite{LINKALIGN} is a framework designed to systematically address schema linking challenges and effectively adapt existing baseline models to real-world environments. The framework comprises three key stages in the form of multi-agent discussion: multi-turn semantic-enhanced retrieval, irrelevant information isolation, and schema extraction enhancement.
\item  \textbf{CHESS:} \textbf{C}ontext-aware, \textbf{H}ierarchical and \textbf{E}xtensible \textbf{S}QL \textbf{S}ynthesizer is an end-to-end text-to-SQL system designed for real-world and complex databases\cite{chess}. It proposes an efficient process centered on a large language model and divided into three major modules: entity and context retrieval, schema selection, and SQL generation. This process can fully utilize database context information and enhance the accuracy and practicality of text-to-SQL in large-scale and heterogeneous schemas.

\item  \textbf{Spider-Agent: } Spider-Agent is a tool-call-based baseline model introduced in Spider 2.0~\cite{spider2.0}. It serves as a crucial benchmark for text-to-SQL tasks on complex databases, facilitating the evaluation of different large language models (LLMs) under unified settings.
\item \textbf{ReFoRCE: } ReFoRCE~\cite{REFORCE} is a leading Text-to-SQL agent on the Spider 2.0 benchmark. It addresses challenges in complex, real-world databases through schema compression, self-refinement, consensus voting, and execution-guided exploration, achieving state-of-the-art results across multiple SQL dialects.
\item  \textbf{MAC-SQL: } MAC-SQL~\cite{MAC-SQL} is a multi-agent Text-to-SQL framework, featuring a Selector for schema selection, a Decomposer for stepwise SQL generation, and a Refiner for query correction based on execution results.
\item  \textbf{TA-SQL: } TA-SQL~\cite{TA-SQL} is a text-to-SQL framework that mitigates schema- and logic-based hallucinations via a Task Alignment strategy, including modules for task-aligned schema linking and logical synthesis, and is evaluated with GPT-4 on the BIRD dataset.
\end{itemize}

\section{More Experiment Results} \label{appendix:more experiment results}

\begin{table}[ht]
\centering
\setlength{\tabcolsep}{1.8mm}{
\begin{tabular}{lccc}
\toprule
Method & SRR & $\bar{C}$ & Avg. Tokens \\
\midrule
MCS-SQL$_{N=1}$& 46.0 & 34.52 & 33.80K \\
MCS-SQL$_{N=2}$ & 54.4 & 39.04 & 67.50K \\
MCS-SQL$_{N=3}$ & 56.0 & 42.86 & 101.50K \\
MCS-SQL$_{N=4}$ & 57.6 & 44.10 & 135.10K \\
MCS-SQL$_{N=5}$ & 58.8 & 45.15 & 168.90K \\
\midrule
SQL-to-Schema$_{N=1}$ & 45.2 & 32.70 & 34.68K \\
SQL-to-Schema$_{N=2}$ & 53.6 & 40.31 & 69.18K \\
SQL-to-Schema$_{N=3}$ & 58.4 & 35.69 & 103.50K \\
SQL-to-Schema$_{N=4}$ & 62.4 & 47.44 & 137.76K \\
SQL-to-Schema$_{N=5}$ & 64.0 & 49.03 & 171.90K \\
\bottomrule
\end{tabular}
}
\caption{Multi-turn results of MCS-SQL and SQL-to-Schema on Spider 2.0-Lite. $N$ represents the times of iterations for decoding.}
\label{tab:saturation-analysis}
\end{table}

\subsection{Limitations of Increasing Decoding Time in Iterative Schema Linking}

Table~\ref{tab:saturation-analysis} reports the strict recall rate, the average number of recalled columns, and the average token consumption for both \textbf{MCS-SQL} and \textbf{SQL-to-Schema} methods as the number of sampling-decoding turns increases on Spider 2.0-Lite. From the results, we observe a clear saturation effect in both strict recall and the number of columns:
At the beginning, increasing the number of $N$ from 1 to 2 or 3 brings a substantial gain in both recall and average recalled columns. However, \textbf{as the number of iterations continues to grow, the improvement becomes marginal—both metrics gradually approach an upper bound}. Specifically, for MCS-SQL, moving from 1 time to 5 times increases SRR from 46.0\% to 58.8\%, and average columns from 34.52 to 45.15; for SQL-to-Schema, SRR increases from 45.2\% to 64.0\%, and average columns from 32.70 to 49.03. Nevertheless, the incremental improvement in the last several rounds is very limited.

\textbf{In contrast, the average token consumption grows almost linearly with the number of $N$,} as more candidates are sampled, decoded, and processed with each additional iteration. This suggests that, although both methods allow one to control the iteration count (N), \textbf{it is difficult to directly control the scale or to reach a recall-quantity comparable to our approach without incurring significantly higher token cost.}

Overall, these findings indicate a practical limitation of simply increasing the number of decoding rounds in such iterative methods: recall and result size eventually saturate, while resource (token) consumption continues to increase, making it challenging to efficiently match the effectiveness and scalability of our proposed approach.

\subsection{Analysis of Hyperparameter}
In this section, we continue to analyze the influence of the hyperparameters \textbf{Max Turn} (The maximum number of dialogue iteration rounds of the agent) and \textbf{top-$m$} in \textbf{@retrieve\_schema}.
\begin{table}[htb]
\centering
\setlength{\tabcolsep}{1.8mm}{
\begin{tabular}{ccccc}
\toprule
Max Turn & Avg. Turn & SRR & Avg. Tokens & $\bar{C}$ \\
\midrule
4  & 3.72 & 89.2 & 20.1K & 138.4 \\
6  & 4.65 & 90.2 & 20.7K & 152.0 \\
8  & 5.01 & 91.0 & 20.8K & 153.5 \\
10 & 5.79 & 91.2 & 21.2K & 159.6 \\
\bottomrule
\end{tabular}
}
\caption{Impact of Max Turn on SRR, Token Cost and Retrieved Columns.}
\label{tab:apx-turns}
\end{table}

\paragraph{Impact of Max Turn.}
Table~\ref{tab:apx-turns} evaluates the effect of varying the maximum number of agent dialogue turns (Max Turn) in \textbf{AutoLink} on SRR, token cost, and average recalled columns. We observe that even with a relatively small Max Turn (e.g., 4 or 6), the method already achieves strong recall (SRR 89.2 and 90.2 respectively) and covers the majority of relevant columns. This demonstrates that \textbf{the agent is able to efficiently explore the schema and discover important elements within the first few interaction rounds.}

As Max Turn increases, both SRR and average recalled columns exhibit only incremental gains—SRR increases by just 2 (from 89.2 to 91.2) and columns by about 21 (from 138.4 to 159.6) as Max Turn moves from 4 to 10, indicating a clear saturation effect. Importantly, the average number of turns taken by the agent grows much more slowly than the allowed maximum (from 3.72 to 5.79 as Max Turn rises from 4 to 10), suggesting that in practice the exploration process tends to converge early, with later rounds contributing diminishing additional gains.

Additionally, the average token cost remains nearly unchanged as Max Turn increases, further confirming that most schema expansion and information acquisition occur in the initial rounds. Overall, these findings indicate that \textbf{AutoLink is capable of rapidly performing effective schema exploration, and strict limits on Max Turn are not necessary to achieve near-optimal recall and coverage.}

\begin{table}[ht]
\centering
\begin{tabular}{cccc}
\toprule
top-$m$ & SRR & Avg. Tokens & $\bar{C}$ \\
\midrule
1 & 89.2 & 20.7K & 152.28 \\
2 & 89.6 & 21.0K & 156.06 \\
3 & 91.2 & 21.2K & 159.40 \\
\bottomrule
\end{tabular}
\caption{Impact of top-$m$ in retrieval on SRR, token cost and retrieved columns.}
\label{tab:apx-topm}
\end{table}

\paragraph{Impact of Top-\textit{m} in Retrieval.}
Table~\ref{tab:apx-topm} examines the impact of the retrieval column number (top-$m$) per \textbf{@retrieve\_schema} action on SRR, token cost, and column coverage. Notably, even with a small top-$m$ setting (e.g., $m$=1 or 2), our method already achieves a high recall rate (SRR of 89.2\% and 89.6\%, respectively) and a substantial number of recalled columns, with little difference compared to larger $m$. As $m$ increases from 1 to 3, SRR improves only slightly (to 91.2\%), accompanied by marginal increases in both average tokens and column counts. This result demonstrates that \textbf{increasing top-$m$ does not lead to significant growth in computational cost or excessive schema expansion.}

The relatively stable performance across different $m$ values indicates that the agent is able to issue precise and targeted semantic queries in \textbf{@retrieve\_schema} steps, retrieving most relevant columns with a small number of candidates per action. Therefore, \textbf{it is not necessary to set a large $m$ value to achieve strong recall, as the agent’s high-quality search queries already ensure efficient coverage of the required schema elements.} This highlights the effectiveness and efficiency of our agent-based schema retrieval framework.
\subsection{Analysis of Trade-off Comparison}
As shown in Figure \ref{trade}, we compare \textbf{AutoLink} (initial schema top-$n$ set to 5, 10, 20, 50, 100), \textbf{MCS-SQL} and \textbf{SQL-To-Schema} (the times of iterations for decoding $N$ set to 1 and 5), \textbf{BGE-Large} and \textbf{BGE-reranker} (retrieval top-$k$ set to 50, 100, 200). Our method AutoLink demonstrates a clear advantage in strict recall rate across varying scales of recalled columns compared to baseline approaches. Specifically, when compared with MCS-SQL and SQL-to-Schema, our method achieves substantially higher strict recall rates. The recall columns returned by MCS-SQL and SQL-to-Schema remain relatively limited, which is primarily constrained by the design of these methods. Both baselines rely on repeatedly decoding with LLMs to extract potentially relevant schema elements. While increasing the number of decoding iterations from 1 to 5 improves both the number of recalled columns and the strict recall rate, the improvement plateaus quickly and is accompanied by a dramatic increase in token consumption. Furthermore, \textbf{these LLM-based methods lack fine-grained control over the recall column size, making a fair comparison with AutoLink under the same recall scale infeasible.}

In contrast, when compared with retrieval-based methods such as BGE-Large and BGE-Reranker, AutoLink consistently achieves higher strict recall rates under the same average number of recalled columns. This highlights the effectiveness of our approach in providing both scalable recall and high precision.
\begin{figure}[!htb]
    \centering
    \includegraphics[width=\linewidth]{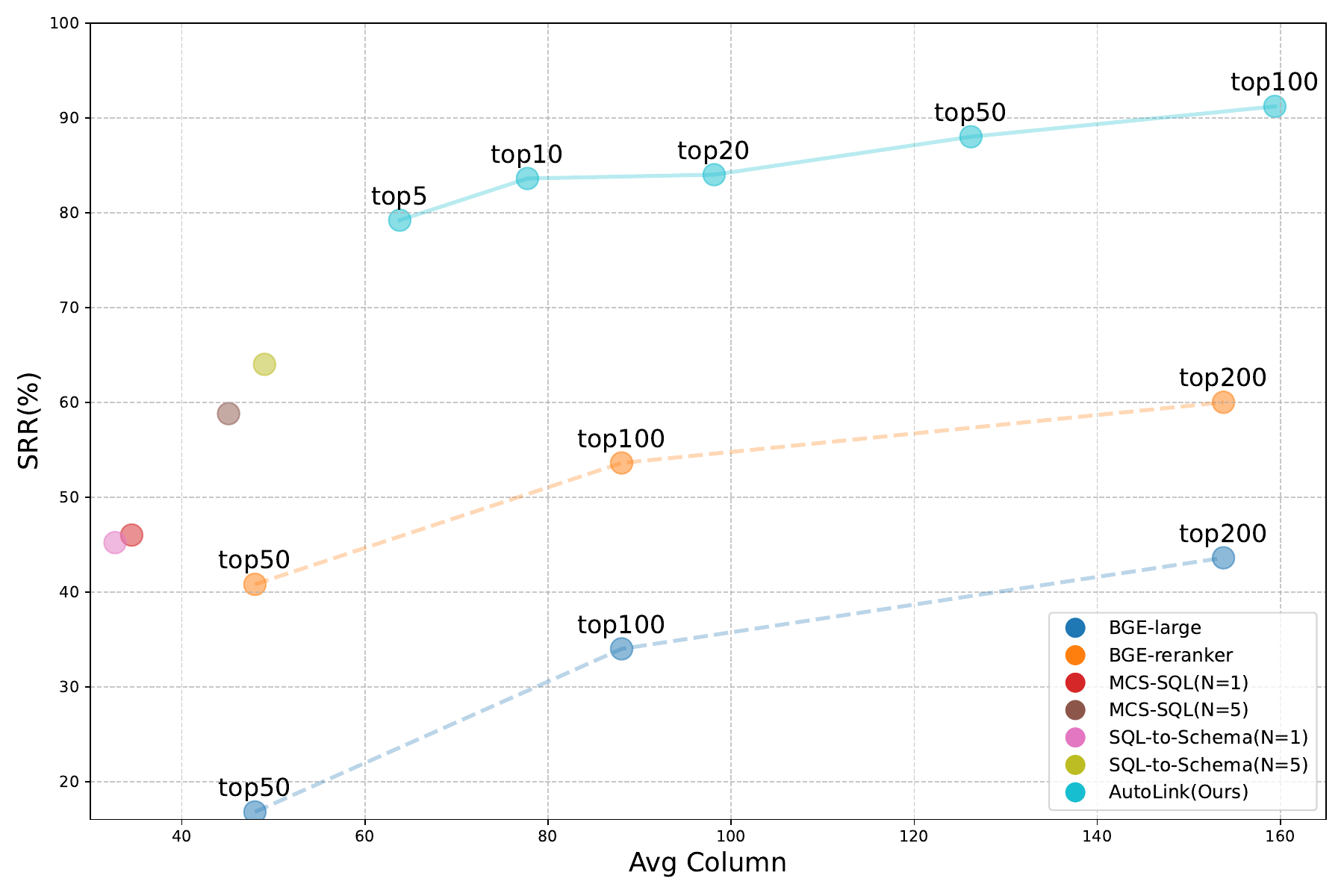}
    \caption{Trade-off comparison between recall columns per instance and strict recall rate across different methods.}
    \label{trade}
\end{figure}

\subsection{Results on Leaderboard of Spider 2.0-Lite}
\begin{figure}[htb]
    \centering
    \includegraphics[width=\linewidth]{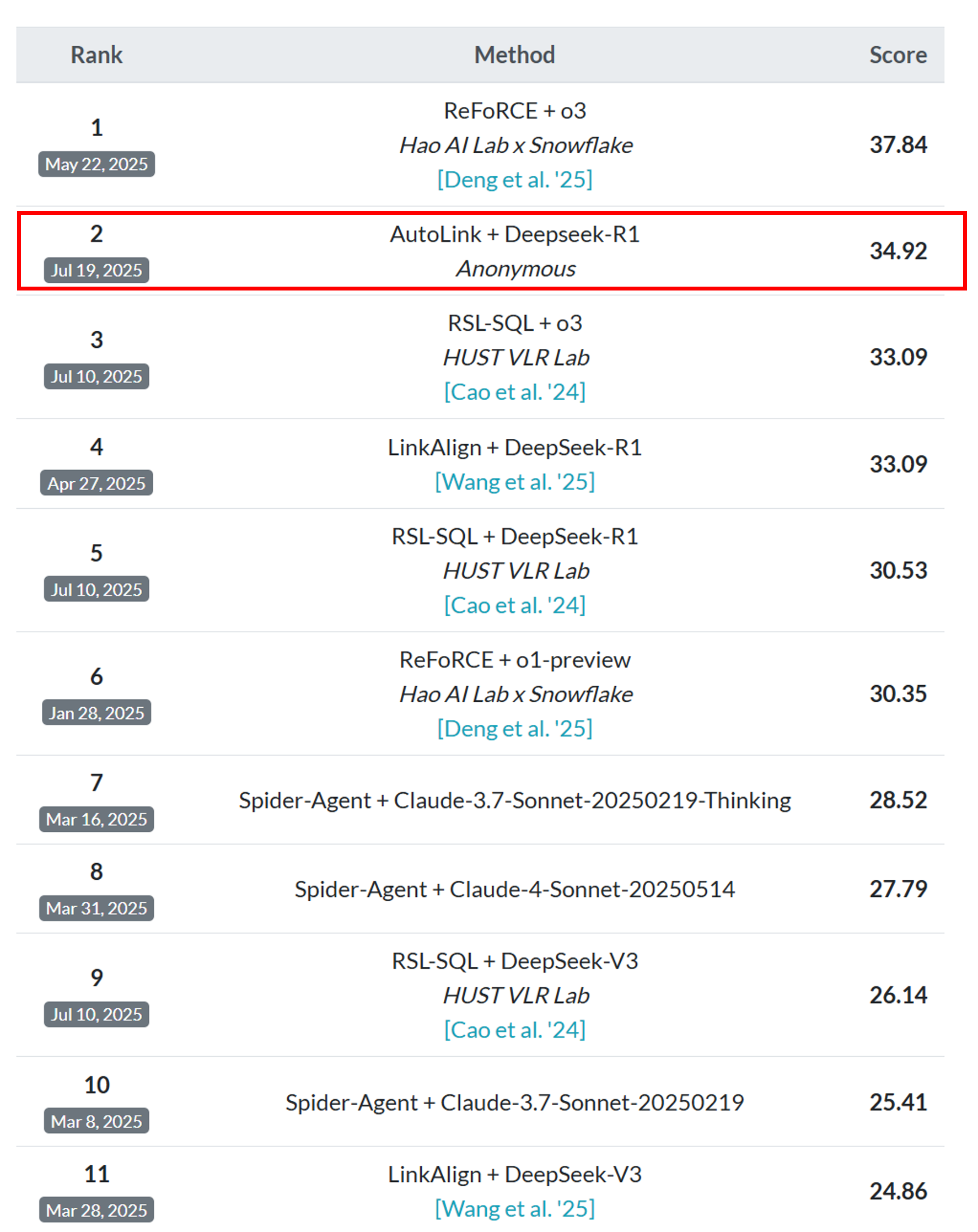}
    \caption{The results on leaderboard of Spider 2.0-Lite. Among all the submitted methods, AutoLink ranked second.}
    \label{lead}
\end{figure}
As shown in Figure \ref{lead}, we present the performance of our method, \textbf{AutoLink}, on the Spider 2.0-Lite leaderboard. At the time of submission, our approach achieves the second-best EX overall. Notably, when compared with other methods utilizing the same DeepSeek-R1 model, AutoLink attains \textbf{state-of-the-art} performance.

\subsection{SQL Generation Ablation}

\begin{table}[ht]
\centering
\begin{tabular}{lcc}  
\toprule
\multicolumn{1}{l}{Method} & \multicolumn{1}{c}{EX}  & \multicolumn{1}{c}{$\Delta$EX} \\ \midrule  
Base Generation  & 23.47  & --  \\
+ Iterative Correction&  31.34 & +7.87 \\
+ Majority Voting  & 34.97  & +3.63 \\ \bottomrule
\end{tabular}
\caption{Ablation for SQL generation on Spider 2.0-Lite.}
\label{ABLATION_gen}
\end{table}
Table~\ref{ABLATION_gen} presents the ablation results for SQL generation on the Spider 2.0-Lite dataset. Starting from the base generation model, which achieves an EX of 23.47, we observe a substantial improvement when iterative correction is applied, raising the EX score by 7.87 points to 31.34. This demonstrates the effectiveness of iterative correction in refining initial predictions and mitigating errors. On Spider 2.0-Lite, due to the complexity of the problem, we found that the model was prone to low-level  syntax errors during the process of generating SQL. Therefore, SQL corrections based on the execution results are necessary. Further incorporating majority voting leads to an additional boost, increasing the EX to 34.97, a gain of 3.63 points over the previous step. These results highlight the complementary benefits of both iterative correction and majority voting, showing that the combination of these strategies significantly enhances the robustness and overall performance of the SQL generation process.
\section{Prompt Templates} \label{Prompt_Templates}
In this section, we present the detailed prompt templates used for the various key modules in our system, namely: (1) Schema Linking in Figure \ref{sl}, (2) SQL Generation in Figure \ref{sg}, (3) Syntactic Correction in Figure \ref{revise}, and (4) SQL Selection in Figure \ref{select}. Each template is designed to guide the Large Language Model (LLM) effectively for its specific subtask.
\begin{figure*}[ht]
    \centering
    \includegraphics[width=0.9\linewidth]{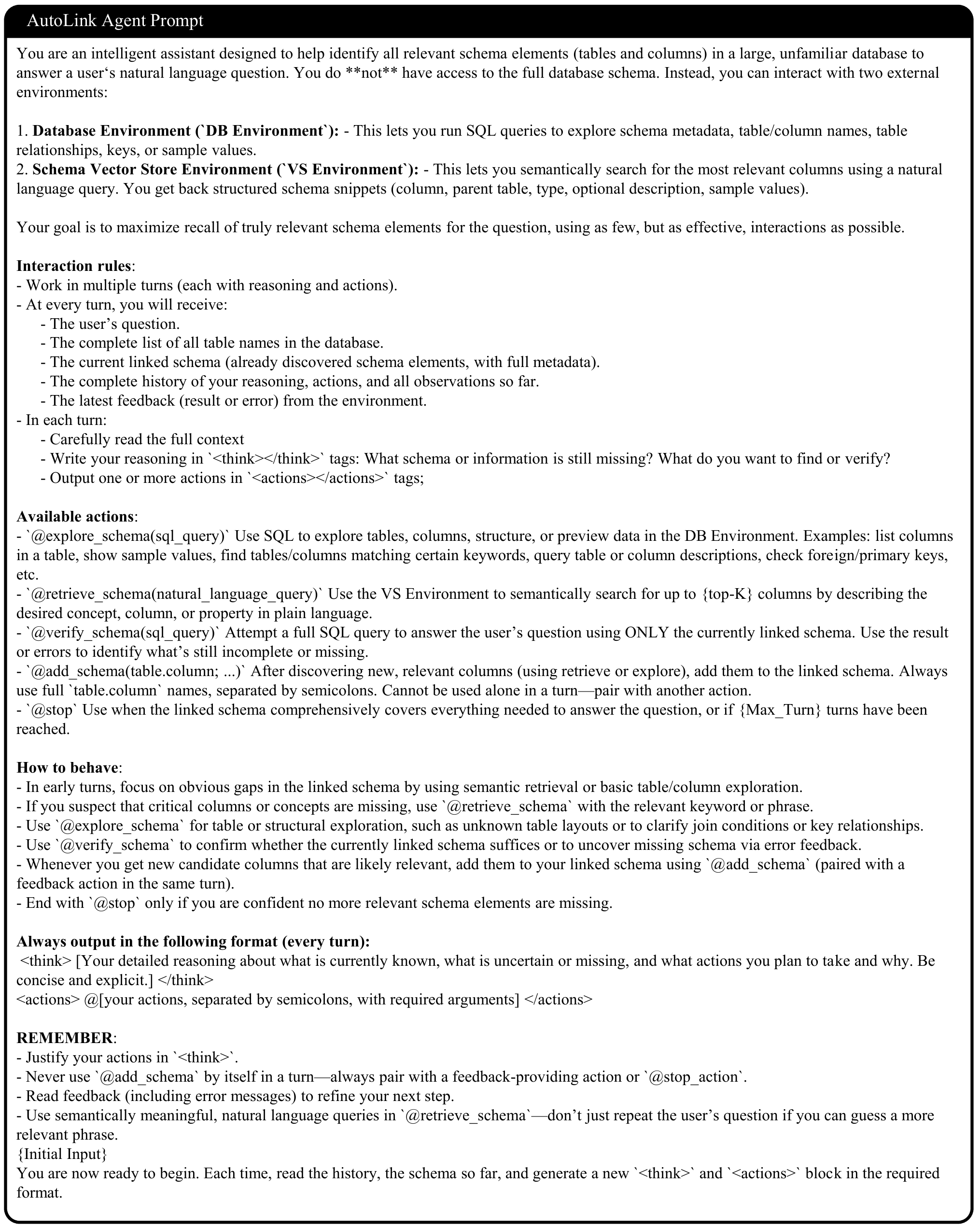}
    \caption{Template of agentic schema linking process.}
    \label{sl}
\end{figure*}
\newpage
\begin{figure*}[ht]
    \centering
    \includegraphics[width=0.9\linewidth]{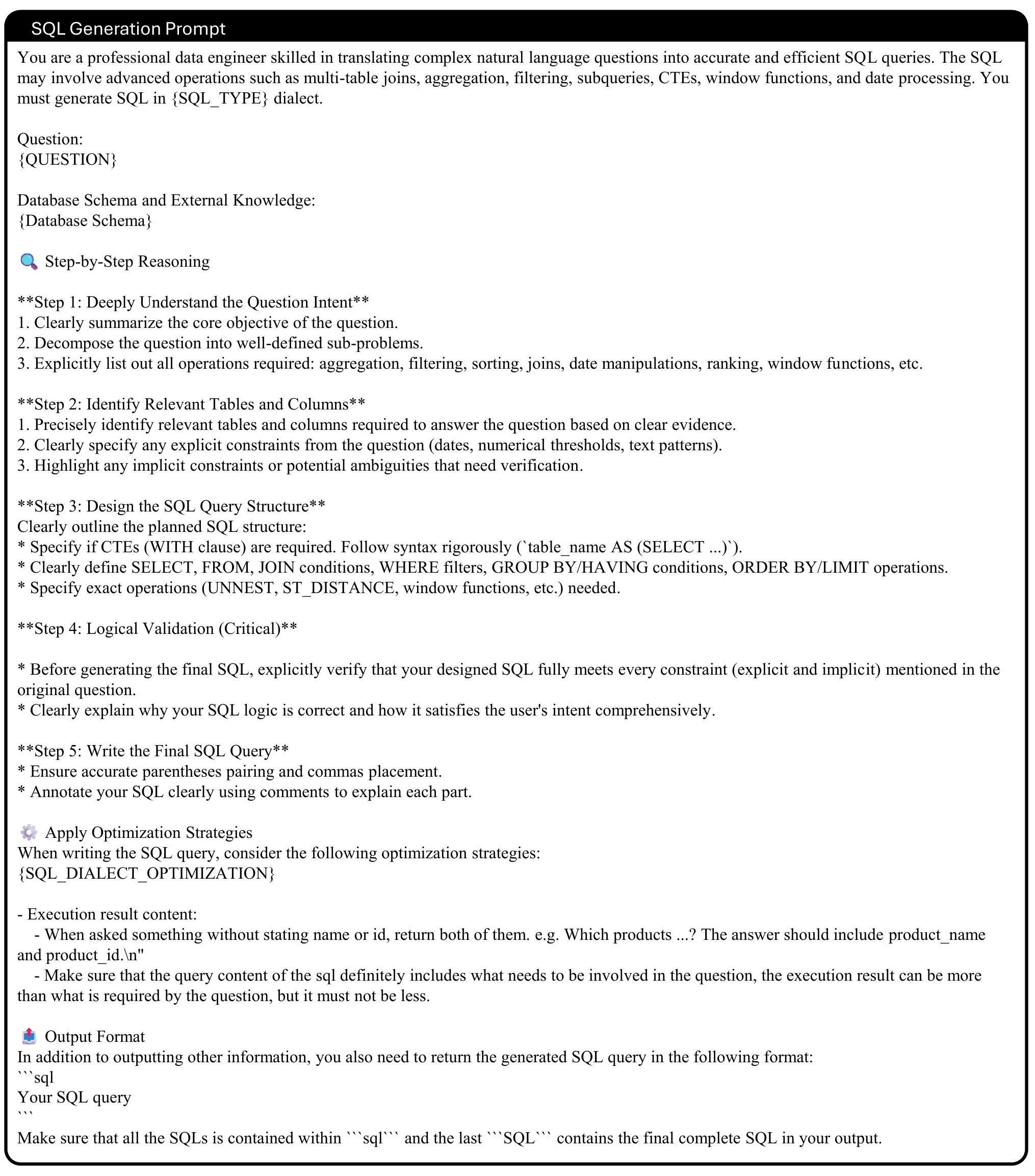}
    \caption{Template of SQL generation.}
    \label{sg}
\end{figure*}
\newpage
\begin{figure*}[ht]
    \centering
    \includegraphics[width=0.9\linewidth]{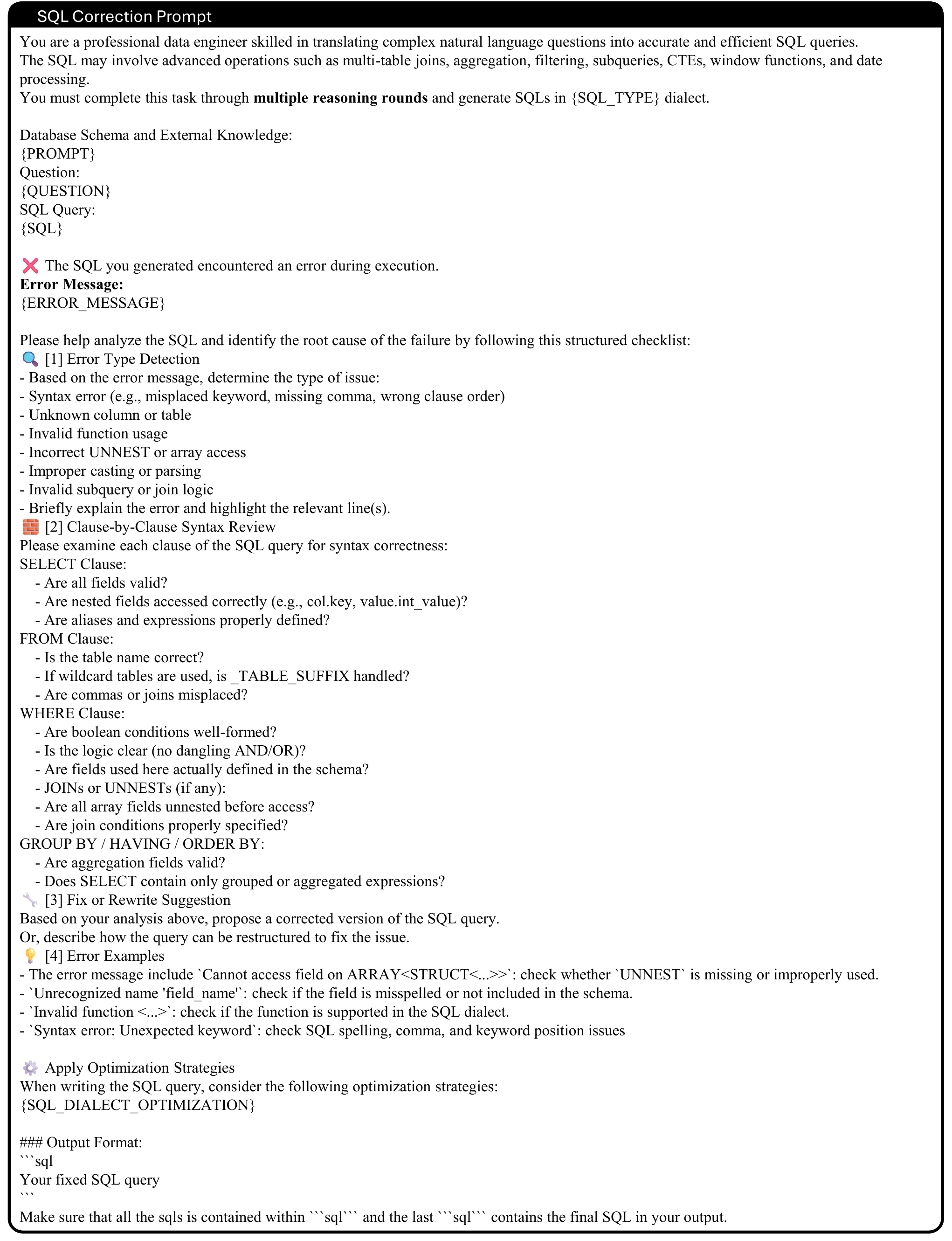}
    \caption{Template of iterative SQL syntactic correction.}
    \label{revise}
\end{figure*}
\newpage
\begin{figure*}[ht]
    \centering
    \includegraphics[width=0.9\linewidth]{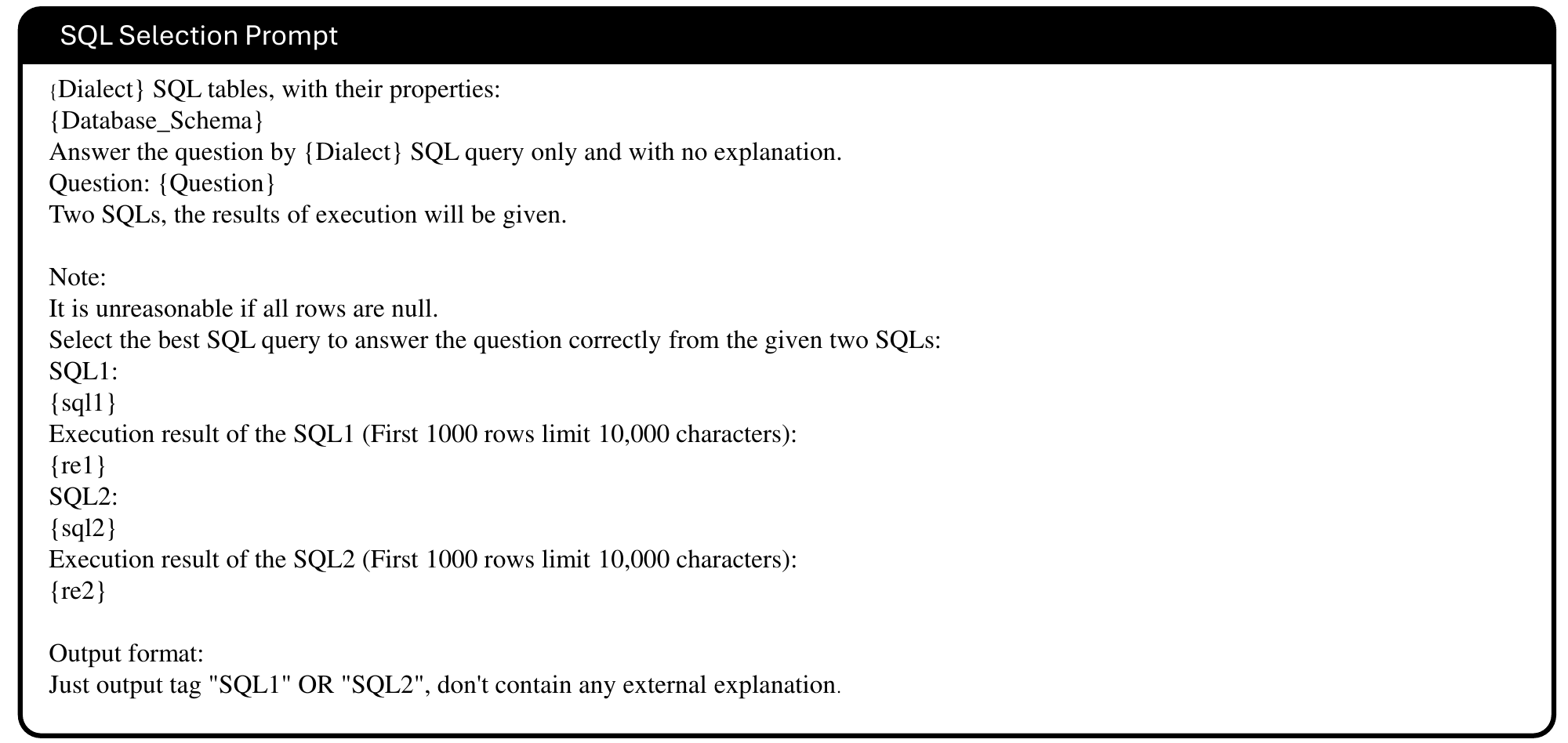}
    \caption{Template of SQL selection.}
    \label{select}
\end{figure*}
\end{document}